\documentclass[11pt]{article}
\usepackage{fullpage}                
\usepackage{hyperref}
\usepackage{graphicx}
\usepackage{amssymb}
\usepackage{epstopdf}
\usepackage{natbib}
\usepackage{linguex}
\usepackage{amsmath}
\usepackage{xcolor}
\usepackage{slashbox}
\usepackage{booktabs}
\usepackage{stmaryrd}
\usepackage{float}
\usepackage{setspace}
\usepackage[title]{appendix}
\usepackage{enumitem}
\usepackage{ulem}

\hypersetup{
	colorlinks   = true, 
	urlcolor     = blue, 
	linkcolor    = blue, 
	citecolor   = blue 
}

\newcommand{\PE}[1]{#1}
\newcommand{\pe}[1]{#1}
\newcommand{\bs}[1]{ #1}

\newcommand{\sem}[1]{\ensuremath{\llbracket #1 \rrbracket}}
\newcommand\posscite[1]{\citeauthor{#1}'s [\citeyear{#1}]}
\newcommand\posscitealt[1]{\citeauthor{#1}'s \citeyear{#1}}

\title{On the Optimality of Vagueness:\\
``Around'',``Between'' and the Gricean Maxims\thanks{Version of July 13, 2022. Acknowledgments to be added.
}}

\author{
Paul \'Egr\'e \and Benjamin Spector \and Ad\`ele Mortier \and Steven Verheyen}

\date{To appear in \emph{Linguistics and Philosophy}}

\begin{document}


\maketitle

\begin{abstract}

\noindent 
Why is ordinary language vague? We argue that in contexts in which a cooperative speaker is not perfectly informed about the world, the use of vague expressions {can} offer an optimal tradeoff between truthfulness (Gricean Quality) and informativeness (Gricean Quantity). Focusing on expressions of approximation such as ``around'', which are semantically vague, we show that they allow the speaker to convey indirect probabilistic information, in a way that can give the listener a more accurate representation of the information available to the speaker than any more precise expression would (intervals of the form ``between''). That is, vague sentences can be \emph{more informative} than their precise counterparts. We give a probabilistic treatment of the interpretation of ``around'', and offer a model for the interpretation and use of ``around''-statements within the Rational Speech Act (RSA) framework. In our account, the shape of the speaker's distribution matters in ways not predicted by the Lexical Uncertainty model standardly used in the RSA framework for vague predicates. We use our approach to draw further lessons concerning the semantic flexibility of vague expressions and their irreducibility to more precise meanings.



\end{abstract}\bigskip


\pagebreak

\section{Introduction}

Why is ordinary language vague? More specifically, what is the function of vagueness in language? Traditional accounts of vagueness generally insist that vagueness is a deficiency, compared to what an ideal language would look like. \cite{russell1923vagueness} defined vagueness as a one-many relation between an expression and its meaning. In an ideal language, the relation would be one-one. And indeed, artificial languages typically eliminate ambiguity and vagueness by the same token.

Various explanations have been proposed to rationalize the vagueness of ordinary language, however. Russell himself made a central observation in connection to language use when he noted: ``\emph{it would be a great mistake to suppose that vague language must be false. On the contrary, a vague belief has a much better chance of being true than a precise one, because there are more possible facts that would verify it}" (p. 91). Phrased in  Gricean terms, Russell's observation may be put as follows: a cooperative speaker who is not perfectly informed about the world would too often flout the Gricean maxim of Quality if compelled to use only precise expressions. The maxim of Quality says that one should not say what one believes to be false, and that one ought not say that for which one lacks adequate evidence (\citealt{grice1967logic}). Conversely, vagueness may help cooperative speakers to remain both truthful and justified in their assertions (see \citealt{egre2018lying}).

The thought, although phrased differently, underlies several accounts or discussions of the use of vagueness, drawing attention to the relation between vagueness and error-reduction in the face of uncertainty (see \citealt{channell1985vagueness, krifka2007approximate,deemter2009utility,lipman2009why, solt2015vagueness}). For Channell, ``vagueness may be used as a safeguard against being later shown to be wrong'' (p. 17). Krifka, relying on earlier remarks by philosopher Pierre Duhem and by anthropologist Elinor Ochs, highlights a tradeoff between precision and certainty, noting that imprecision can be a form of prudence in conversational exchanges. Similarly, van Deemter points out that ``the doctor is uncertain how the future will turn out,
which is why he, sensibly, wraps his predictions in vagueness'' (p. 622), \PE{and Solt emphasizes that vagueness ``reduces speaker's commitment'' (p. 123), creating an advantage in promissory situations exemplified in political and legal contexts. Lipman himself, after arguing that vagueness poses a challenge for a game-theoretic account of communication, notes that vagueness offers a way of accommodating ``unforeseen contingencies'', leaving open whether this could help rationalize vagueness.\footnote{\pe{\cite{lipman2009why}, in a signaling-game setting, proves that using a vague language, defined as using a mixed strategy over signals, cannot be optimal compared to using a pure strategy, which Lipman interprets as using a precise language. Prima facie, this result may seem to go counter to the result of optimality we establish in this paper. However, Lipman's definition of vagueness in fact departs significantly from ours, \bs{and we do not endorse it}. We leave a comparison between his result and our approach for another occasion.}}}


{The idea that the function of vague language \PE{could be} to help speakers comply with the maxim of Quality is \PE{both natural and plausible. But it does not suffice to explain vagueness}, given the availability of semantically precise but logically weak sentences that make it easy to satisfy the maxim of Quality even in cases where one has little information. For instance, a sentence such as \emph{There were between 2 and 97  guests at the party yesterday} has intuitively precise truth-conditions but expresses a logically weak statement, which can be used truthfully by a speaker who knows fairly little. In contrast, an utterance of a vague sentence such as \emph{There were about 45 guests at the party yesterday} intuitively suggests that the speaker has more information than one who utters the former precise sentence.

In what follows, we will argue that certain vague expressions allow for an optimal tradeoff between the maxims of Quality and Quantity. They sometimes allow the speaker to achieve a communicative effect that no semantically precise sentence could. 
More specifically, focusing on expressions of approximation such as ``around'', we will argue that such expressions allow the speaker to indirectly convey probabilistic information, so as to comply with the maxim of Quality while achieving high informativity. {Where  previous work emphasized how vague language can function as a safeguard against potential violations of truthfulness, we claim that vague sentences can help speakers maximize informativity.}

The probabilistic dimension of the interpretation of vague expressions has antecedents in the literature.  \cite{frazee2010vagueness} argue that
 gradable  terms like ``tall'' or ``many'' are vague in so far as they constrain ``some measure
relative to a value which cannot be known in principle or in practice''. On their approach, and in agreement with standard theories of the context-sensitivity of gradable expressions, ``tall'' semantically means ``taller than $t$'', and ``many'' means ``more than $m$'', but speaker and hearer are typically uncertain about those threshold values $t$ and $m$ (see \citealt{barker2002dynamics,kennedy2007vagueness}). Frazee and Beaver's picture of communication, which we endorse in this paper, is that the ``information conveyed by a vague sentence is a statistical distribution'' over values and thresholds, which interlocutors try to convey to each other. They argue that the use of vague language is rational in situations of uncertainty, and our own proposal is, \pe{in} this respect, close in spirit to theirs. \citet{lassiter2017adjectival} offer a probabilistic model of the pragmatics of vague predicates within the Rational Speech Act model of pragmatics (\citealt{goodman2013knowledge}). In their model, when interpreting a sentence such as \emph{she is tall}, the listener updates her belief state (viewed as a probability distribution about possible states of the world) in a way that factors in uncertainty about thresholds. The communicative effect of the sentence can then be viewed as the way it affects this posterior belief state. {While the model we develop in this paper is to a significant extent inspired by \posscite{lassiter2017adjectival} work, it is worth pointing out that \posscite{lassiter2017adjectival} paper focuses on situations where the speaker is } maximally informed about the variable of interest (say, someone's height), and so does not by itself address the link between vagueness and speaker's uncertainty ({in section  \ref{sec:comparison} we discuss in more details the Lexical Uncertainty model of \citealt{bergen2012s,bergen2016pragmatic}, which is closely related to  \posscitealt{lassiter2017adjectival} model, and provides one possible way of extending it to the general case where the speaker is not fully informed}).}

	\bs{In this paper, we focus on the meaning, use and interpretation of  expressions of numerical approximation such as ``around'' and ``about''. We present a model where such vague sentences end up communicating a probability distribution. More specifically, the speaker, though not fully informed, is assumed to have more information than the listener about some variable of interest, and the sentence used is informative to the extent that the listener's posterior distribution over world states after processing an utterance is closer to that of the speaker than prior to the utterance}.  We argue that the reason why speakers may choose a vague statement as opposed to a precise (but logically weak) one is that this allows them to achieve some communicative effects that would not have been achievable by using a precise statement. In particular, coming back to Russell's intuition, vague language might allow speakers to be both informative and truthful even when their epistemic state does not categorically rule out any particular state of affairs, by allowing them to indirectly convey that they take some state of affairs to be more likely than others.

 Our first goal is to identify a range of contexts in which the use of ``around'' is optimal compared to any lexical alternative that would be more precise (section \ref{sec:motivations}). Our second goal is to advance the understanding of the probabilistic semantics of ``around'' by giving specific attention to the comparison between numerical expressions of the form ``around $n$'' and the use of precise intervals of the form ``between $i$ and $j$'' (sections \ref{sec:around} and \ref{sec:ratio}). The basic treatment we give of ``around'' in section \ref{sec:around} is fundamentally listener-oriented. In section \ref{sec:speaker}, \bs{we provide a model of how the speaker chooses her messages, and explain the sense in which an ``around''-message can be more informative than any competing ``between''-message}. \bs{In our model, the choice between an ``around''-message and a ``between''-message can depend on the \emph{shape} of the speaker's probability distribution over the variable of interest: the ``around-n''-message is preferred by a speaker whose distribution favors values close to $n$.} Sections \ref{RSAmodel} -\ref{concrete} explain how to integrate this model within \bs{the Rational Speech Act framework of \citet{goodman2013knowledge} (RSA for short).} In section \ref{sec:comparison}, we propose a detailed comparison with the model of gradable adjectives proposed by \cite{lassiter2017adjectival}{, and the Lexical Uncertainty Model of \cite{bergen2016pragmatic}. \bs{Readers not interested in the technical details of the various RSA models we discuss can skip sections \ref{RSAmodel}-\ref{sec:comparison} and} go directly to section \ref{sec:further}, where we outline some limits of the current model and further potential developments. \bs{Section \ref{sec:flex} then discusses how \pe{our account {positions} itself in the debate between epistemic and semantic accounts of vagueness (see \citealt{sorensen1988blindspots,williamson1994vagueness, wright1995epistemic}), and argues that vague meanings are irreducible to precise meanings.} \pe{Last, section \ref{sec:conclusion} recapitulates our main findings in this paper.} 
 
 \pe{The paper includes three technical appendices}. \pe{Appendix \ref{app:LU}  \bs{ proves a result discussed in section \ref{sec:comparison}, namely: in the Lexical Uncertainty Model, in contrast with our model, the speaker's choice of a message only depends on the \emph{support}, not the \emph{shape} of the speaker's distribution over the variable of interest.}} 
\bs{{Appendix \ref{app:original} describes a variant of our 
model, originally our first model, making fundamentally identical qualitative predictions, but distinct quantitative predictions. Appendix \ref{app:variants} presents two alternative models briefly discussed in section \ref{sec:comparison}.}}}


\section{When vague is better than precise}\label{sec:motivations}

When is it rational for a cooperative speaker to use vague as opposed to more precise language? 

One class of situations concerns cases in which the speaker is fully knowledgeable and has precise information at her disposal. She may prefer to use vague language, however, if she expects a precise figure to convey irrelevant information to the listener. For instance, to use an example from \cite{veltman2001verschil}, suppose the question under discussion is how fast you can run the steeplechase. The speaker may prefer to say ``I can run the steeplechase very fast'' than to utter ``I can run the steeplechase in 11 minutes 12 seconds'', if she expects the listener to not have the slightest idea of racing times in relation to the steeplechase. By using the vague predicate ``very fast'', the listener can get more efficient information about the speaker's relative position compared to other runners than if absolute temporal information were communicated.\footnote{Interlocutors typically assess vague expressions relative to implicit standards. See for example \cite{verheyen2018subjectivity} for empirical evidence that ``tall'' and ``heavy'', applied to human figures, are ascribed in part relative to one's own height and weight. 
\pe{However,  Veltman's effect can be produced by using an exact but a proportional/relative expression of comparison {(e.g., ``I am in the top 7\%''')}. 
It cannot be said, therefore, that vague expressions are \emph{necessary } in order to communicate relative information.} \pe{We are indebted to A. Cremers 
and to an anonymous reviewer} for this observation.} {Similarly, a fully knowledgeable speaker may choose to use an `around'-statement with a round number, when the conversational context does not require her to provide precise information. For instance, I may inform you that I have ``around 30 students in my class'' when I know that I have exactly 29 students (we do not focus on such uses in this paper, but we briefly return to this point in section \ref{sec:rounding}).\footnote{See \cite{henst2002truthfulness} for a discussion of related cases, in which speakers prefer to round off the time even when they know it with high precision. Explicit use of ``around'' is not needed in such cases, ``thirty students'' can be used to mean ``around thirty students'', or ``five o'clock" to mean ``about five o'clock", as discussed by \cite{lasersohn1999pragmatic}.}}

A second class of situations, which will be our main focus in this paper, concerns cases in which the speaker herself fails to have precise information at her disposal. Such cases loom large in \cite{williamson1994vagueness}'s epistemic account of vagueness, and they are described by \cite{deemter2009utility} as cases of necessary vagueness. One of Williamson's central examples concerns a subject watching a crowd, and unable to make an exact count of the people in the crowd. Suppose the speaker is attending a party involving a crowd of 77 people, but does not know that number. Imagine that, after the party, the speaker is asked: ``how many people were at the party?''. By assumption there is no number $n$ for which the speaker can respond: ``there were exactly $n$ persons at the party'', on pain of violating Grice's maxim of Quality. For either the speaker would fail to say something she thinks is true, or she may by luck give the correct answer, but she would fail to have adequate evidence for it.\footnote{\pe{This relies on both parts of Grice's maxim of Quality, including ``do not say that for which you lack adequate evidence'', which can be strengthened into \cite{williamson2000knowledge}'s norm of assertion (``assert only what you know'').}}

How then should the speaker respond? Our account is grounded in the assumption that situations of that kind fundamentally involve a probabilistic representation of the situation. In practice, the speaker has a probability distribution on possible values of the number of invitees at the party. Let us assume, for the sake of the argument, that the speaker knows with certainty that there were no more than 100 invitees, and knows with certainty that there were at least 40 people present. By assumption, the support of the speaker's probability distribution \bs{(i.e. the set of values which are assigned a non-null probability)} is the interval $[40, 100]$. We may suppose moreover that her distribution has a peak at 70, and that 80\% of the probability mass lies between the values 60 and 80 on either side of that peak (see Figure~\ref{fig:crowd}). \bs{From a normative point of view, distributions with such a `peaked' shape are expected to arise whenever one receives a \emph{noisy} signal from a source, and one knows the signal to be noisy (for instance the signal received -- say a number -- is sampled from a Gaussian distribution centered on the  `true' number).}

\begin{figure}
\caption{Hypothetical probability distribution regarding the number of people present at a party.}\label{fig:crowd}
\[
\includegraphics[scale=.4]{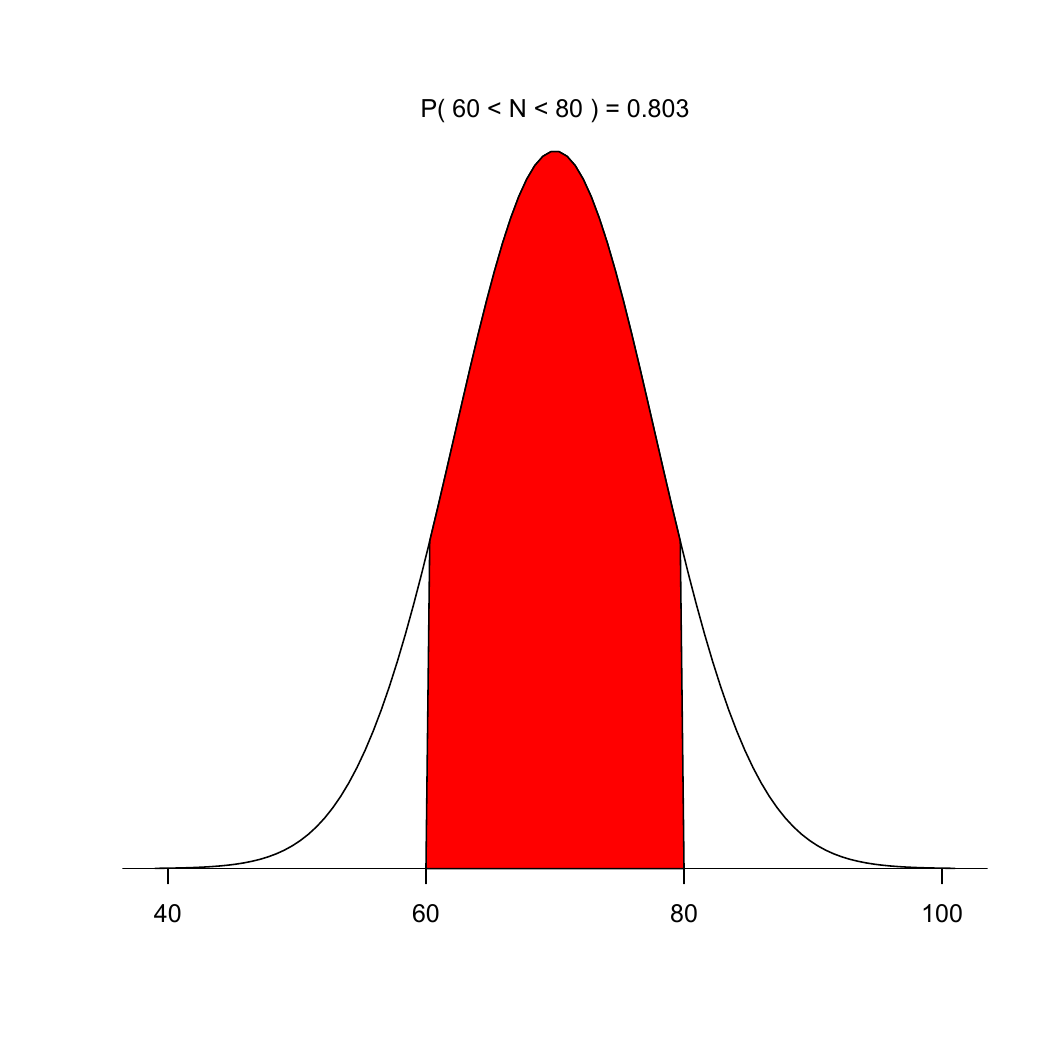}
\]
\end{figure}
 
Assuming the speaker has access to her distribution, the speaker has several strategies in response to the question. \PE{Let us consider three of them, which have in common that they do not use vague vocabulary}. The first is to communicate the support of her distribution. By uttering ``there were between 40 and 100 persons" the speaker is guaranteed to satisfy the maxim of Quality. Intuitively, however, the speaker reports poorly on her information state. Although the proposition ``there were between 40 and 100 persons at the party'' is the most informative proposition she believes with certainty in response to the question, the information communicated gives no hint concerning the values the speaker deems more likely than others.

 A second option would be for the speaker to communicate a narrower interval. For example, she may choose to respond ``between 60 and 80'', if she thinks that a .8 chance of getting the right result is high enough. This time, the speaker communicates a more informative proposition, still likely to include the true value. Nevertheless, in case 85 people were at the party, the speaker runs the risk of ruling out the actual value and of misleading the listener. A case of this kind is a potential violation of the maxim of Quality, or of Williamson's knowledge norm. 
 
 
 A third option would be for the speaker to communicate her confidence level: ``I believe with 80\% confidence that there were between 60 and 80 persons at the party''. However, that proposition is complex to articulate. Moreover, it forces the speaker to say something explicit about her epistemic state.\footnote{Accessing one's confidence level may be a delicate matter too. One might argue that even reports of confidence levels are challenges to the Knowledge norm of assertion. However, our argument above does not rely on that premise, we may assume that the speaker has reliable access to her confidence levels. \PE{In practice the speaker can also say ``probably between 60 and 80 persons'', or ``I believe between 60 and 80 persons'', but they involve vague ``hedges'' (in the sense of \cite{lakoff1973hedges}).}} 
  
\PE{There is a fourth option, however, which is to use vague vocabulary.} In particular, by uttering ``there were around 70 people at the party'', the speaker is avoiding the pitfalls of each of the previous options. First of all, even if 85 or 55 people turned up, the speaker is not ruling out either possibility by saying ``around 70''. In that sense, the utterance is safe and able to respect Quality. Secondly, ``around 70'' intuitively conveys information about the shape of the speaker's distribution: 70 should be taken to be more probable to the speaker than other values, and so this message conveys probabilistic information that the ``between-sentence'' does not. Thirdly, by saying ``around 70'' the speaker does not have to communicate an explicit confidence interval.\footnote{\PE{As pointed out by a referee, empirical sciences contain reports of magnitude estimates of form $193\pm5$ cm. This indicates that the true value lies in the interval $(188, 198)$ with a certain probability, and it assumes a specific distribution (usually Gaussian) centered on 193. ``Around''-sentences of natural language specify neither the distribution nor the boundaries, but our main point will be that they communicate that the target value is comparatively more probable than more remote values.}}
In what follows, we propose to substantiate those intuitions. In order to do so, we proceed to specify a model of the interpretation of ``around'' in the next section.

\section{Modeling ``around''}\label{sec:around}

\subsection{Around vs. Between} ``Around'' is sometimes interpreted as specifying a fixed interval whose extension depends on the granularity of a contextually given measurement scale. According to \cite{krifka2007approximate} and \cite{solt2014alternative}, ``around $n$'' denotes the interval $[n-\frac{u}{2}, n+\frac{u}{2}]$, where $u$ is the unit setting the relevant granularity. For instance, ``around 10'' could denote the interval $[9,11]$, or $[9.5,10.5]$, or $[5,15]$, and so forth, depending on the context. 

We agree that the meaning of ``around $n$'' should be cashed out in terms of intervals centered on $n$, but we believe it is inadequate to specify this meaning in terms of a unique fixed interval. Even as the granularity is known to all speakers, say is equal to 10, ``around 20'' need not be interpreted rigidly as meaning ``between 15 and 25''. To wit, a speaker who fails to know how many people were at a party and to whom intervals of ten units set the order of magnitude would not necessarily speak falsely by reporting ``around 20'' if the actual number of attendees were 26 or 27.

We thus see two main differences in the comparison of ``around'' and ``between''. First of all, ``around'' is \emph{semantically vague}, whereas ``between'' is not. This means that ``around $n$'' does not specify a sharp interval; it is compatible with an open-ended range of values, unlike ``between''. Consider the following two reports:

\ex.\label{around} There were around 70 people at the party.

\ex.\label{between} There were between 60 and 80 people at the party.

Intuitively, if we learn that there were 87 people at the party, \ref{between} appears false \emph{stricto sensu}, unlike \ref{around}. Of course, the utterer of \ref{between} may be using ``between'' with some slack, and a charitable listener may deem \ref{between} close enough to the truth to be acceptable. However, the point is that for ``around'' the vagueness in question is directly part of the meaning of the expression. Further confirmation of the vagueness of ``around'' is given by modification of the target numerals with ``exactly''. This modification is permitted with ``between'' but produces gibber with ``around'':

\ex. \a. ??There were around exactly 70 people.
\b. There were between exactly 60 and exactly 80 people.

Relatedly, it can be observed that ``around $n$'' is sorites-susceptible in a way that ``between $i$ and $j$'' isn't. For us, this means that if ``$k$ is around $n$'' is considered true, then ``$k'$ is around $n$'' is also likely to be judged true when $k'$ is close enough to $k$ but a little more removed from $n$ (\citealt{tcs, egre2019sorites}). For example, if 19 is around 30, then it seems that 18 is also around 30. But if 20 is between 20 and 30, 19 is not between 20 and 30. For ``between'' we thus expect the membership function to be a step function, but for ``around'' we expect a smooth function.\footnote{On sorites-susceptibility and the need for smooth membership functions, see also \cite{borel:1907,smith2008vagueness, egre2014borel}.} 

The second main difference we see when comparing ``around $n$'' and ``between $i$ and $j$'' is more subtle, but will occupy central stage in the rest of this paper. We call it \emph{peakedness}. It concerns the representation of how probable the values are in the interval $[i,j]$ specified by ``between'', compared to those in neighbourhoods of $n$ in the case of ``around''. Assume you have no idea how many people, within a certain range, will attend the next evening lecture at the university. You ask the organizer how many people she expects. Compare the following answers:

\ex. \a. Between 20 and 40.
\b. Around 30.

Our intuition is that \Last-b conveys that the closer a value is to 30, the more likely it is deemed by the speaker in this case. In particular, \Last-b conveys that 30 is more likely to the speaker than other values. By contrast, \Last-a does not appear to convey that any value in the range $[20,40]$ is more likely than any other: no peakedness results in this case.

The intuition in question is subtle here. We think it will be particularly clear in contexts where, before processing the sentence, the listener does not have strong expectations as to how many people will turn up. This will translate into the assumption that the listener has a uniform prior on the number of attendees, within a certain range. When the prior is not uniform, so when some values are initially more expected than others, peakedness remains in play, as we discuss in section \ref{sec:ratio}, but it may be less manifest.


\subsection{A semantics for ``around''} \label{sec:semaround}

In order to derive the previous facts, we propose a Bayesian model of the interpretation of ``around''. {The model is actually a variant of a distinct model that we first came up with and that will be presented later (\PE{see Appendix \ref{app:original})}. The two models agree in their main predictions, but an advantage of the Bayesian model is that its conceptual motivation is very clear.

The model has two components: a \emph{semantic } component which specifies the meaning of \emph{around}, and an \emph{inferential component } which describes how listeners update their beliefs, when they accept a sentence, on the basis of its meaning. Importantly, the model we propose in this section is listener-oriented. That is, we first account for the effect that using ``around'' is producing on the listener. We consider the speaker's perspective in section \ref{sec:speaker}.

\subsubsection{The model} We assume, following the spirit of a number of former proposals, the truth-conditional meaning of ``$x$ is around $n$'' is that $x\in A^n_y$, where $A^n_y$ is an interval of the form $[n-y, n+y]$ and {$y$ is an open semantic parameter:

\ex. \sem{\text{around}}$^y$ = $\lambda n. \lambda x. x \in [n-y, n+y]$

From a purely semantic point of view, then, an utterance of the form ``$x$ is around $n$'', cannot express a proposition unless a value for $y$ is provided. However, even if no specific value for $y$ is provided, the listener nevertheless learns something from such an utterance, namely the fact that, whatever the value of $y$ is, $x$ in in the interval $[n-y, n+y]$. If the listener has some expectations about the values that  $y$ could take, then she can gain information regarding $x$. This is, in essence, the idea that our model of the listener will capture. That is, the listener's task is to infer what values $x$ is likely to have, given some uncertainty on what values $y$ is also likely to have. The key point will be that, under uncertainty about the length of the intended interval, a value closer to $n$ is more likely to fall in that intended interval. In this respect, our proposal is close in spirit to  \posscite{lassiter2017adjectival} approach to gradable adjectives: the taller Mary is, the more likely it is that her height is above the treshold for \emph{tall}, so when learning that Mary is tall, the listener shifts her probability distribution over Mary's height to higher values.\footnote{{One significant difference is that in \posscite{lassiter2017adjectival} proposal, which is couched in the Rational Speech Act model, the joint reasoning about the variable of interest -- say someone's height -- and the parameter of interpretation (e.g., a threshold for \emph{tall}) does not take place at the level of the `literal listener', but is carried out by the first-level pragmatic listener. This difference will prove to have important consequences when we develop a model for the speaker. See section \ref{sec:comparison} and Appendix \ref{app:LU} for a detailed discussion.} \label{fnlassiter1}}

We represent the listener's information state by a joint probability distribution $P$ over the possible values taken by $x$ and $y$. $P(x=k)$ represents the prior probability that $x$ takes on a specific value $k$, and $P(y=i)$ represents the prior probability that the interval picked by ``around'' has radius $i$. We assume that the prior probabilities of these two types of events are independent, so that in general:
\begin{align*} P(x=k, y=i) = P(x=k)\times P(y=i)\end{align*}
Let $n$ be the number used in `$x$ is around $n$'. {The information gained by the listener is that $x$ is in the interval $[n-y, n+y]$, where both $x$ and $y$ are random variables. We assume that the goal of the listener is to infer the correct value of $x$}.
{That is, the listener's problem is to figure out the conditional probability distribution defined by 

\begin{equation*}
P(x=k\mid \text{$x$ is around $n$}) = P(x=k\mid x \in [n-y, n+y])
\end{equation*}

We have:}

\begin{equation*}\label{eq:bay_around}
P(x=k\mid \text{$x$ is around $n$})=\sum\limits_i P(x=k, y=i \mid x \in [n-y, n+y]) \end{equation*}
\noindent 
Let us abbreviate $d(x,n)\leq y$ for $x \in [n-y, n+y]$, which is equivalent to $|n-x|\leq y$. {Bayes Theorem allows us to rewrite  each term of the sum in the preceding equation as follows:}

\begin{equation*}
P(x=k, y=i\mid d(x,n)\leq y) = P(d(x,n)\leq y\mid x=k, y=i)\times \dfrac{P(x=k, y=i)}{P(d(x,n)\leq y)}\end{equation*}
Note that $P(d(x,n)\leq y\mid x=k, y=i)$ equals $1$ if $d(k,n)\leq i$, and is $0$ otherwise. Let $\mathcal{I}$ be defined as a function which takes an arithmetic statement and returns its truth-value. Then we have:
\begin{equation*}P(x=k, y=i\mid d(x,n)\leq y) =\dfrac{{\mathcal I}(d(k,n)\leq i) \times P(x=k, y=i)}{P(d(x,n)\leq y)}\end{equation*}
The denominator does not depend on either $k$ or $i$. Let us call it $D$ to ease calculations.\footnote{$D = \sum\limits_k P(x=k)\times \sum\limits_{i \geq |n-k|} P(y=i)$ --- this is the sum of all terms that can be obtained from the numerator by instantiating $x$ with all its possible values.} We therefore have:
\begin{align*}\label{eq:around}
P(x=k\mid \text{$x$ is around $n$}) & =\sum\limits_i \dfrac{{\mathcal I}(d(k,n) \leq i) \times P(x=k, y=i)}{D} \\
&= \dfrac{1}{D}\sum\limits_{i\geq|n-k|} P(x=k)\times P(y=i)\nonumber \\
& = \dfrac{1}{D} \times P(x=k) \times \sum\limits_{i \geq |n-k|} P(y=i)
\end{align*}
In the remaining of this paper, we will often use the \emph{proportionality notation}, whereby the above equation is expressed as follows (\PE{we name it BIR, for Bayesian Interpretation Rule}):%
\footnote{The symbol $\propto$ reads `is proportional to'. More specifically, a statement of the form $f(a| \ldots) \propto g(a,\ldots)$ is shorthand for: $f(a|\ldots) = \frac{g(a,\ldots)}{\sum\limits_{a' \in A}g(a',\ldots)}$, where  $A$ is the domain over which the variable $a$ ranges. So the above formula 
boils down to: 
	
$$P(x=k\mid \text{$x$ is around $n$}) = \dfrac{P(x=k) \times \sum\limits_{i \geq |n-k|} P(y=i)}{\sum\limits_{\text{$j$ is in the support of $x$}} P(x=j) \times \sum\limits_{i \geq |n-j|} P(y=i)}$$.

The proportionality factor ensures that the sum of all $P(x= \ldots \mid \text{$x$ is around $n$})$, across all possible values for $x$, is $1$, so that $P(x=\ldots \mid \text{$x$ is around $n$})$ is a probability distribution. \label{prop}}

\begin{equation}\label{eq:around2}
P(x=k\mid \text{$x$ is around $n$}) \propto  P(x=k) \times \sum\limits_{i \geq |n-k|} P(y=i)\nonumber \tag{\textcolor{blue}{BIR}}
\end{equation}

\subsubsection{Illustration} \label{ill}

To illustrate the predictions of this model, let us assume that the listener's prior distribution is uniform both on the values that $x$ might have, and on the interval values $y$ that ``around'' might denote. Upon hearing ``around $n$'', we may assume that $[0,2n]$ is the largest interval compatible with the meaning of ``around $n$'' (if $0$ is the scale minimum), hence that $[0,n]$ is the range of values that $y$ can take.\footnote{
{Given that we want $n$ to be the middle of the interval, making $[0,2n]$ the largest possible interval is natural when the numeral is used to talk about the cardinality of a set - since negative numbers are then not relevant. \cite{ferson2015natural} provide experimental data showing that when asked to estimate the largest interval compatible with an \emph{around n}-statement, people tend to pick an interval that is much narrower than $[0,2n]$. Taken at face value, this could suggest that the prior distribution on $y$  should categorically exclude too large intervals -- since otherwise the posterior distribution resulting from an ``around''-sentence would not categorically }{exclude any value that was not already excluded prior to the utterance. Such a conclusion is however not warranted, and depends on the `linking' theory that provides the bridge between a specific model and people's behavior in an experimental task. In our setup, even with a uniform prior distribution on the set of intervals of the form $[n- i, n+i]$, with $i \leq n$, as well as on the range of the variable of interest $x$, the listener's posterior distribution after processing ``around $n$'' assigns very low probability to values that are very far from $n$. It is very plausible that, when asked to estimate an interval, people simply report an interval of values which receive a high enough probability, and therefore exclude values which, without being equal to 0, are in practice negligible.}}

For such uniform priors, it follows analytically from Equation \eqref{eq:around2} that:

\begin{equation*}\label{eqaround}
P(x=k \mid \text{$x$ is around $n$}) = \dfrac{n-|n-k|+1}{(n+1)^2} 
\end{equation*}

Consider for instance the effect of a Listener hearing ``$x$ is around 20''. The posterior distribution of the Listener on the possible values of $x$ is depicted by the histogram (black lines) in Figure \ref{fig:around20}. The red line represents the uniform prior on the values $x$ might take. As the figure makes clear:\medskip

\begin{figure}[t]
\caption{In black: posterior probability of $x=k$ for ``around 20'' from uniform priors on values $x$ and interval radii $y$; in red \PE{(lower line)}: uniform prior on $x$; in blue \PE{(upper line)}: posterior for ``between 10 and 30''.}\label{fig:around20}
\[
\includegraphics[scale=0.4]{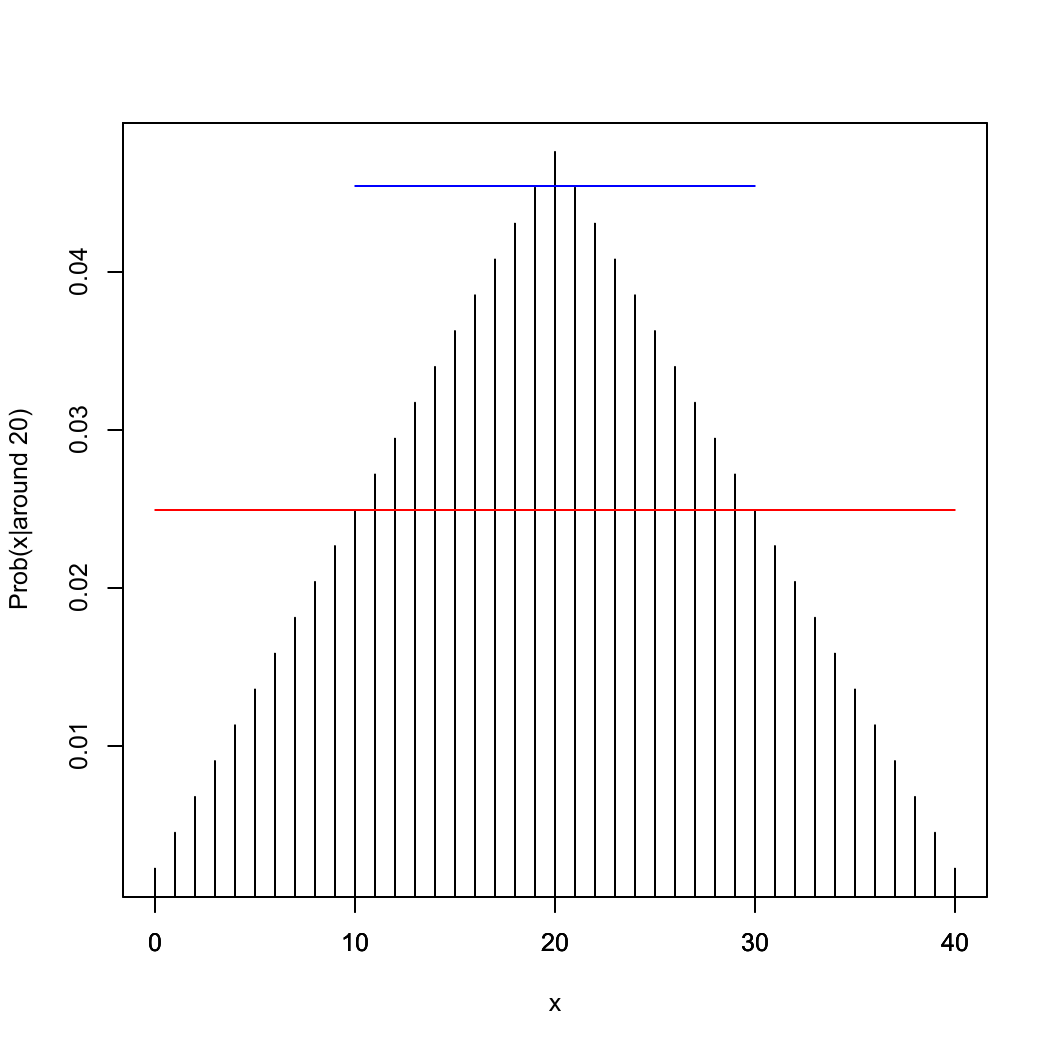}
\]
\end{figure}

(i) For ``$x$ is around 20'' the posterior distribution is symmetric and centered on 20, and the further away a value is from 20, the less probable it has become.\medskip

(ii) Compare hearing ``$x$ is between $10$ and $30$'. In the latter case, the Listener does a simpler Bayesian update assigning zero probability to the values outside the interval $[10, 30]$. The solid blue line represents the posterior for ``between 10 and 30'', which is a uniform posterior.\medskip

\subsubsection{Main features} From this example, we see that the model captures the two main differences pointed out earlier concerning ``around'' and ``between''. 

First of all, the vagueness of ``around'' is represented: semantically, the meaning of ``around'' does not specify a fixed interval. Moreover, in this example the posterior obtained for ``around'' captures sorites-susceptibility: we see that close values are assigned close posterior probabilities and that the posterior is a smooth function. For ``between'', by contrast, the meaning is crisp, and the posterior is given by a step function involving a ``jolt'' between some consecutive values (\citealt{smith2008vagueness}). 

Secondly, peakedness is captured: starting from a flat prior, ``around'' outputs a nonuniform posterior in which values closer to the target number are more probable, whereas ``between'' outputs a flat posterior.

Finally, we note that in this example, the idea that the vagueness of ``around'' is a safeguard against error for the speaker is also present. From the listener's perspective, it is compatible with hearing ``around 20'' to assign a nonzero probability to the value 0, if indeed 0 was an option to her initially. Of course, the listener's posterior on 0 is very small and negligible in comparison to other values, but a speaker who would choose to specify a sharp interval using ``between'' incurs a higher risk of error in a situation in which they may suspect the listener to not rule out any value initially.\footnote{We return to this point, and to its philosophical significance, in section \ref{sec:conclusion}.} 

\section{The ratio inequality}\label{sec:ratio}

In the previous example, it was assumed \PE{for the sake of illustration} that the priors were uniform, but \emph{this need not be the case in general}. Suppose $P(x=k)$ is non-uniform to begin with. Then using ``between $i$ and $j$'' will rescale that nonuniform prior to a new nonuniform posterior. Conversely, if values other than $n$ are deemed initially more probable, the posterior may not show a peak on $n$ for ``around $n$''. \PE{Crucially, however, our arguments in this paper make no assumption about the priors attached to either ``between'' or ``around''}: that is, we can still show that the effect of using ``around'', compared to using ``between'', is to increase the probability of `central' values relative to less central ones, by assigning more weight to numbers nearer the target value. 

This fact is captured by the following ratio inequality, which says that when a number $k_1$ is closer to the target value $n$ than some other number $k_2$, then the ratio of the posterior probabilities of $k_1$ and $k_2$ upon hearing `` $x$ is around $n$'' is larger than the ratio of their priors. 
More formally, let $k_1$ and $k_2$ be two integers such that $k_1 < k_2$, then: 

\begin{equation*}
\dfrac{P(x=n-k_1 \mid \text{$x$ is around $n$})}{P(x=n-k_2 \mid \text{$x$ is around $n$})} > \dfrac{P(x=n-k_1)}{P(x=n-k_2)}
\end{equation*}\medskip

Let us show this. Call the first ratio (on the left-hand side of the above inequality) $R_{\textrm{post}}$ and the second one $R_{\textrm{prior}}$. From Equation \eqref{eq:around2}, it follows that:

\begin{align*}
R_{\textrm{post}} & = \dfrac{P(x= n-k_1)\times \sum\nolimits_{i \geq k_1} P(y=i)}{P(x=n-k_2)\times\sum\nolimits_{i \geq k_2}  P(y=i)}\\
 & = \dfrac{P(x=n-k_1)}{P(x=n-k_2)}\times \dfrac{\sum\nolimits_{i \geq k_1} P(y=i)}{\sum\nolimits_{i \geq k_2} P(y=i)}\\
 & =R_{\textrm{prior}}\times \dfrac{\sum\nolimits_{y = k_1}^{y=k_2-1} P(y=i) + \sum\nolimits_{y \geq k_2} P(y=i) }{\sum\nolimits_{y \geq k_2} P(y=i)}\\
 & \PE{\textrm{{\footnotesize (since $k_2>k_1$, by decomposition of the numerator on the right)}}}\\
 & {= R_{\textrm{prior}} \times (1 + \dfrac{\sum\nolimits_{y = k_1}^{y=k_2-1} P(y=i)}{\sum\nolimits_{y \geq k_2} P(y=i)})  > R_{\textrm{prior}}}
\end{align*}

From the ratio inequality, it follows that the ratio of the posteriors for ``around'' is greater than the ratio of the posteriors for ``between'', since for ``between'', the ratio of the posteriors must be equal to the ratio of the priors. That is, let $k_1$ and $k_2$ belong to the interval $[i, j]$ specified by ``between $i$ and $j$'', then:

$$\frac{P(x=k_1\mid \textrm{between $i$ and $j$})}{P(x=k_2\mid\textrm{between $i$ and $j$})}=\frac{P(x=k_1)}{P(x=k_2)}$$\medskip

As a result, the model predicts that even if the priors on possible values of $x$ and on interval values of $y$ are nonuniform to begin with, using ``around $n$'' instead of ``between $i$ and $j$'' (for $i\leq n \leq j$)  will give $n$ a comparatively higher posterior. The ratio inequality matters, because it makes a prediction regarding the relationship between the posterior distributions generated by ``between" and ``around" statements which is independent of the prior probability distribution. This prediction can in principle be investigated empirically.\footnote{We refer to \cite{mortier2019master} 
for a report on some preliminary results. \bs{In this study, we asked participants to select an interval for an \emph{around n} statement, and then asked them to report weights for each value in the selected interval, both for the \emph{around}-sentence and for a \emph{between}-sentence based on the same interval. The results confirmed the prediction made by the ratio inequality, but further work is needed to control for potential confounds.}}

\section{A model of the speaker}\label{sec:speaker}

Now that we have a precise model of the listener, we can provide a model for the speaker which will capture the idea that a speaker may prefer a vague `around'-sentence over a precise `between'-sentence if the `around'-sentence leads the listener to a posterior probability distribution which is closer to that of the speaker.

In most approaches to pragmatics in formal semantics and philosophy of language, informativity is defined in terms of \emph{entailment}, and information states are modeled as sets of possible worlds, i.e. propositions. Grice's maxim of quantity is then interpreted as a requirement that the speaker communicate the proposition that matches her information state. Consider a speaker who knows that the value of $x$ is in a certain interval $[a,b]$, and does not know anything else. Then, if asked about the value of $x$, the best the speaker can do is to say something like ``$x$ is between $a$ and $b$''. In particular, an ``around''-sentence fails to categorically rule out values that are outside of $[a,b]$, and so should never be preferred. 

In our probabilistic setting, information states are more fine-grained, and are represented as probability distributions over states of the world (`worlds' for short). Imagine, in particular, that at some point the listener and the speaker had exactly the same information about some variable of interest (say the number of people who attended a certain party). Then the speaker makes a private observation that brings her to a new information state (a new probability distribution over the variable of interest). Let us assume that this private observation cannot 
be directly communicated. The goal of the speaker is then, in our setting, to find a message such that, when the listener will process it, the posterior probability distribution of the listener over the variable of interest will be as close as possible to hers. 

What we need, therefore, is to specify the underlying {\emph{measure} that the speaker will use in order to assess how close to her own distribution the posterior distribution of the listener will be after processing her message}. The {measure} we use comes from information theory, and is known as \emph{Kullback-Leibler Divergence}, in the wake of recent decision-theoretic models of pragmatics (esp. models couched in the \emph{Rational Speech Act} framework, see \citealt{frank2009informative,goodman2013knowledge,bergen2016pragmatic}). Before giving the formula for this {measure}, we first motivate it, and introduce the relevant information-theoretic background. Readers familiar with the concept of K-L divergence may go directly to section~\ref{speaker choice}; others may find this a fruitful preamble, as the notion is often taken for granted.\footnote{For a general introduction to K-L divergence, see \citet[179]{mcelreath2016rethinking}. The presentation we give is more specific to the communicative framework under discussion.}

\subsection{Information, surprisal, and the Kullback-Leibler Divergence}\label{sec:KL}

Assume that the speaker and the listener start with the same prior probability distribution $P$ over world-states, which we simply identify to the possible values of some variable of interest, noted $x$. Then the speaker makes a private observation $o$, as a result of which she has a  new probability distribution over world-states, notated $P_o$. When the speaker uses a message $m$ (for instance `$x$ is around 7', or `$x$ is between 5 and 9'), the listener processes it and ends up with a posterior distribution $P_m$ (for instance, if $m$ is `$x$ is around $n$', we have $P_m(x=k) = P(x = k | x \textrm{ is around } n)$, which we computed above).

Now, suppose after these two events happen, the state of the world, that is some number $k$, is observed by both the speaker and the listener. The more \emph{unlikely} the observed number was relative to their probability distributions, the more surprised they are, and the more information they get. In information theory, the information gained by an agent when observing $k$ is equated to $-\log(P(x = k))$, where $P$ is the probability distribution that represents the agent's epistemic state before the observation.\footnote{\PE{We use the natural logarithm throughout this paper.}}

\ex. 
\a. Listener's surprisal when observing $k$, after having processed $m$:\\ $-\log(P_m(x=k))$
\b. Speaker's surprisal when observing $k$, after having observed $o$:\\ $-\log(P_o(x=k))$

Since the listener may fail, after hearing the message, to fully recover the information that the speaker has, the listener whose probability distribution is $P_m$ would be, on average, more surprised, when learning the true state of the world, than the speaker whose probability distribution is $P_o$ (that is, observing the true state of the world world brings more information to someone who is not very knowledgeable than to someone who is more knowledgeable). Now, saying that the speaker wants to bring the listener to a state as close as possible to hers amounts, in this setting, to the idea that the speaker would like to minimize, across worlds, the difference in {future} surprisal between the listener and herself (ideally the listener would fully recover the information that the speaker has, and this average difference will be 0, in which case, if both observed the true state of the world, they would both be exactly as surprised).

\ex. Difference of surprisals between Listener and Speaker after observing $x=k$:\\
$-\log(P_m(x=k)) - (- \log(P_o(x=k))) = \log(P_o(x=k)) - \log(P_m(x=k))$

The speaker does not know which world is in fact the case, so she wants to minimize the `average', or `expected' difference in surprisal between her and the listener in case they observed the actual world. But whose expectations should we use to compute this expected difference in surprisal?  Suppose that there are only two worlds $w_1$ and $w_2$, and $P_o(w_1) = 0.9$ and $P_o(w_2) = 0.1$, while $P_m(w_1) = P_m(w_2) = 0.5$. Upon observing $w_2$, the speaker would be more surprised than the listener, so in this case the difference in surprisal between Listener and Speaker would be negative. The speaker has good reasons to think that $w_2$ is in fact very unlikely. It is much more likely for $w_1$ to be observed, and in this case the listener would be more surprised than the speaker.  
And this second possibility should receive more weight, since in fact, given the additional information that the speaker has, it is more likely to occur. That is, because the speaker's probability distribution results from a truthful observation, and therefore corresponds to a better epistemic state than that of the listener, the \emph{expected} difference in surprisal between speaker and listener should be computed \emph{from the perspective of the speaker}, and will be the following weighted average:

\ex. $P_o(w_1)\times(\textrm{\emph{difference in surprisal between Listener and Speaker if }} w_1\textrm{\emph{ is observed}})$\\$ + P_o(w_2)\times(\textrm{\emph{difference in surprisal between Listener and Speaker if }} w_2\textrm{\emph{ is observed}})$

Generalizing from this simple case, we get the following:

\ex. Expected difference in surprisal between Listener and Speaker, from the point of view of the speaker who has observed $o$:
\begin{align*}
& \sum\limits_{k}P_o(x=k) \times [\log(P_o(x=k)) - \log(P_m(x=k))]\\
&=\sum\limits_{k}P_o(x=k) \times \log \left(\dfrac{P_o(x=k)}{P_m(x=k)} \right)
\end{align*}

This quantity is known as the \emph{Kullback-Leibler divergence} of $P_m$ from $P_o$.\footnote{For any two probability distributions $P_1$ and $P_2$, $D_{KL}(P_1||P_2)$ is always positive. This reflects the fact that it measures the gain in information when one starts with a distribution $P_2$ and makes an observation which results into a posterior distribution $P_1$. In case $P_1$ cannot be rationally reached from $P_2$ (because $P_2$ assigns probability 0 to some world-states that are assigned a non-null probability by $P_1$), the KL-divergence is infinite (see footnote \ref{quality}).\label{information gain}}

\ex. $D_{KL}(P_o||P_m) = \sum\limits_{k}P_o(x=k) \times \log \left(\dfrac{P_o(x=k)}{P_m(x=k)}\right)$

\subsection{The speaker's utility function and choice rule} \label{speaker choice}

The goal of a cooperative speaker who has observed $o$ will be to pick a message $m$ that \emph{minimizes} the quantity we have just defined. To capture this idea, we can define a \emph{utility} function which defines the \emph{payoff} that the speaker gets from using message $m$ if her information state if $P_o$, i.e. if she observed $o$, such that this payoff \emph{increases} as $D_{KL}(P_o||P_m))$ decreases.\footnote{It is worth mentioning here that using this measure indirectly captures Grice's maxim of Quality. This is for the following reason. Suppose that the speaker is in no position to exclude a certain world-state $x=j$, i.e. $P_o(x=j) > 0$. Suppose she picked a message that would in fact exclude this state, i.e. such that $P_m(x=j) = 0$. Then the quantity $log\dfrac{P_o(x=k)}{P_m(x=k)}$ can be viewed as infinite (because the denominator in the fraction is 0), and so one term of the sum in the formula above will be infinite, as a result of which $D_{KL}(P_o||P_m))$ is infinite, and $U(m,o)$ is infinitely negative. \label{quality}}

\ex. $U(m,o) = - D_{KL}(P_o||P_m)$

We now assume that, when making a choice between several messages $m_1, m_2, \ldots$,  a speaker who has observed $o$ picks the message $m_i$ such that for every $j \neq i$, $U(m_i, o) > U(m_j, o)$.\footnote{For simplicity, we ignore the possibility that two messages are exactly tied, i.e. have exactly the same utility. Furthermore, we assume here that the speaker is fully rational and picks the best message with probability 1. In Rational Speech Act models, the speaker is typically not assumed to be fully rational. Rather, she follows a so-called  `SoftMax'-rule whereby she is more likely to use the best message than the second best, more likely to use the second best than the third best, etc., but nevertheless does not pick the best message with probability 1. This difference is not important at this point. However in section \ref{RSAmodel}, where we develop a fully explicit RSA model, we use the SoftMax rule. \label{FullRationality}}

\subsection*{A case where the speaker prefers an `around'-statement}

We now provide a description of a case where a speaker would, according to our model, receive a higher payoff from using an ``around''-statement than from using a ``between statement''.  Our goal is to provide an existence proof, that is to show that our model makes it possible for an ``around''-sentence  to be a better message than any ``between''-sentence.

We assume the value of interest $x$ can range from 0 to 8, and that initially both the speaker and the listener have a uniform distribution over $x$ (for any $k, P(k) = \frac{1}{9}$). \PE{Again, this assumption of uniform priors is only for the sake of simplicity, and is made without loss of generality regarding our argument}. Then the speaker makes an observation as a result of which she has a new probability distribution over worlds-states, notated $P_o$, which categorically excludes only the values 0 and 8, but is extremely biased towards central values, giving a 96\% probability to the interval $[3,5]$ (cf. Table \ref{Po}).

\begin{table}[t] 
\caption{Hypothetical Speaker's distribution $P_o$.\label{Po}}\medskip
\centering
\begin{tabular}{r|ccccccccc}
  \hline
$k$ & 0 & 1 & 2 & 3 & 4 & 5 & 6 & 7 & 8 \\ 
\hline
  $P_o(k)$ & 0 & 0.01 & 0.01 & 0.16 & 0.64 & 0.16 & 0.01 & 0.01 & 0 \\ 
   \hline
\end{tabular}
\end{table}

It is clear that the optimal `between'-message for the speaker is `$x$ is between 1 and 7', since this message exactly specifies the support of the speaker's distribution. But the speaker could also use the message `$x$ is around 4'. Now, we consider the posterior probability distributions of the listener after processing both messages (Table \ref{Pm}), {assuming the listener has a uniform prior on the intervals expressed by ``around''}. After processing the `between'-message, the listener ends up with a uniform distribution on the interval $[1,7]$. After the `around'-message, using the equation in Section \ref{ill}, the listener ends up with a probability distribution that does not categorically rule out any value, but gives more weight to central values. 

\begin{table}[t]
\caption{Listener's posterior distributions after hearing  `between 1 and 7' and `around 4'.\label{Pm}}\medskip
\centering
\begin{tabular}{r|ccccccccc}
  \hline
$k$ & 0 & 1 & 2 & 3 & 4 & 5 & 6 & 7 & 8 \\ 
\hline
 $P_\textrm{\emph{between}}(k)$ & 0 & 0.14 & 0.14 & 0.14 & 0.14 &  0.14 &  0.14 &  0.14 & 0 \\
  $P_\textrm{\emph{around}}(k)$ & 0.04 & 0.08 & 0.12 & 0.16 & 0.20 & 0.16 & 0.12 & 0.08 & 0.04 \\ 
   \hline
\end{tabular}
\end{table}

While neither of these distributions is intuitively close to the hypothesized speaker's distribution, the one that results from the \emph{around}-message is biased towards central values, like the hypothesized speaker distribution. On the other hand, it fails to exclude the values $0$ and $1$, which the speaker's distribution excludes. When we now compute the KL-divergence of each of these distributions from the speaker's distribution, it turns out that we get a smaller value with $P_\textrm{\emph{around}}(k)$ than with $P_\textrm{\emph{between}}(k)$, hence the ``around''-sentence is better at reducing the distance between the listener's distribution and the speaker's distribution.\footnote{The reason this is the case is that for every value $k$ in $[1,7]$, $P_{\textrm{\emph{around}}}(k)$  is closer to $P_o(k)$ than $P_{\textrm{\emph{between}}}(k)$ is. $P_{\textrm{\emph{between}}}$ wins only for the extreme values $0$ and $8$ but since these values have anyway a null probability of occurrence according to $P_o$, they do not play any role in the computation of the expected difference in surprisal, which is computed from the point of view of the speaker.} We have $D_{KL}(P_o||P_\textrm{\emph{between}}) = 0.89$, while $D_{KL}(P_o||P_\textrm{\emph{around}}) = 0.65$. This translates into the following speaker utilities for each message: $U$(`$x$ is around 4', $o$) = -$0.65$ and $U$(`$x$ is between 1 and 7', $o$) = -$0.89$. The speaker will thus receive a higher utility from the ``around''-message, and is therefore predicted to use it.\footnote{\cite{parikh1994vagueness} contains a pioneering account of the informational value of using vague sentences, which also consists in exhibiting a context in which the listener benefits from hearing a vague term. However, his argument is not the same. His point is basically that when speaker and listener have different semantic representations for a vague term, provided those representations overlap sufficiently, using that term can communicate more information to the listener than \emph{not} using it, given some purpose (for instance, saying ``the book is titled ``X'' and it is blue'', instead of just ``the book is titled ``X''"). Our argument is different, and makes a stronger claim, for we compare the informational value of using a vague term to the informational value of using \emph{precise counterparts} for that term (rather than that of not saying anything).}

\bs{The prediction that the speaker's choice between an \emph{around}-message and a \emph{between}-message is influenced by the specific shape of her subjective distribution can lend itself to empirical testing. For now, we can consider the following thought experiment. Suppose that a certain playground is open to children who are at least four years old and at most ten years old, and that both Isabel and Nassim know this. Isabel has never seen little Sarah, and all she knows is that Sarah is currently playing in the playground. Nassim also knows that little Sarah is currently playing in the playground, but on top of that, and unlike Isabel, he could also briefly glance at her. They are then both asked what they know about Sarah's age. One of the two says `She is between 4 and 10 years old', and the other says `She is around 7 years old'. If someone familiar with the situation is asked who said what, we predict that the \emph{between}-sentence will typically be attributed to Isabel,  and the \emph{around}-sentence to Nassim. Specifically, Nassim, in this scenario, has more information than Isabel. Both know that little Sarah is between 4 and 10 years old, and neither can categorically exclude any specific value in this interval, but on top of that Nassim was able to form an imprecise estimate of Sarah's age, which would typically result in a peaked distribution.}

\section{A full RSA-model for `around'} \label{RSAmodel}

So far, our model of the listener does not take into account the fact that the speaker chooses her message strategically (as just discussed), and that the listener can use her knowledge of the speaker's choice rule to derive additional inferences. Rather, our listener, when interpreting an ``around $n$''-message, simply conditionalizes her joint probability distribution on $x$ (the variable of interest) and $y$ (which determines the length of the interval corresponding to an ``around''-statement) with the information that $x$ is in the interval $[n  - y, n+y]$, but does not derive any inference about the speaker's epistemic state.

Now, a \emph{pragmatic} listener might reason that if the speaker picked an ``around''-sentence rather than a ``between"-sentence, this might precisely be because the speaker's probability distribution is biased towards central values, so that the ``around''-sentence was a better sentence to use than one based on ``between''. Such an extra inference might then strengthen the conclusion that central values are more likely than peripheral values. And an even more sophisticated speaker might take this into account when choosing her message, making the ``around''-message even more appropriate when the speaker's epistemic state is biased towards central values. 

In this section, we provide a model which can capture these extra inferences. This model is a particular version of the Rational Speech Act framework. It first defines a listener of level-0 who is just the one we have defined in section \ref{sec:semaround}. This listener simply conditionalizes her joint distribution on $x$ and $y$ with the information carried by an ``around''-sentence in virtue of its linguistic meaning, namely the proposition that $x$ is the interval $[n  - y, n+y]$, where both $x$ and $y$ are variables whose values are not known.\footnote{Importantly, {as noted in footnote \ref{fnlassiter1}}, we substantially depart from \posscite{lassiter2017adjectival} model and, more generally from models with lexical uncertainty (e.g., \citealt{bergen2012s,bergen2016pragmatic}), which we  discuss in section \ref{sec:comparison} and Appendix \ref{app:LU}. In such models, the literal listener is relativized to a fixed value for $y$, and there are as many literal listeners as they are possibly values for $y$. It is only at the level of the first \emph{pragmatic } listener that uncertainty about the value of $y$ is factored in.} Importantly, this basic listener draws no inference about the epistemic state of the speaker. Then a first-level pragmatic speaker is defined along the lines of the speaker model introduced in section \ref{sec:speaker}. But then we can define a pragmatic listener (called the `first-level pragmatic listener', $L^1$ for short) who knows that she received a message from the first-level pragmatic speaker, and who updates (using Bayes' rule) her probability distribution over both the variable of interest $x$, and a variable $o$ ranging over the possible epistemic states of the speaker. On this basis, we can then define a second-level pragmatic speaker who chooses her message strategically, still with the goal of making the listener's epistemic state about $x$ (the variable of interest) as close as possible to hers, based on the assumption that the listener she is speaking to is the first-level pragmatic listener. This process can continue indefinitely, and defines an infinite sequence of listeners and speakers -- the higher we are in the sequence, the more pragmatically sophisticated the speaker and the hearer are.

We now proceed to describe such a model in detail. We first describe it `in the abstract' and then present a specific implementation. Our goal is to show how the basic effect we have been discussing can get amplified through pragmatic reasoning.

\subsection{Set-up}\label{setup}

We assume that the  listener cares about the value of some variable $x$ which ranges over the natural numbers between $0$ and $8$. Before the speaker makes a private observation, they share a joint probability distribution $P$ over pairs $\langle x,o\rangle$, where $x$ is the variable of interest, and $o$ ranges over a set of \emph{observations} $O$ that the speaker could in principle make (we will specify the set of observations as well as other important ingredients of the model in Section \ref{concrete}). Then the speaker makes a private observation $o_j$, as a result of which her new probability distribution is $P_{o_j}$, defined by $P_{o_j}(x = k) = P(x=k\ |\ o = o_j)$. Furthermore, the listener has a prior probability distribution over the variable $y$, which determines the interpretation of \emph{around}, as discussed above. We assume that $y$ is probabilistically independent of $x$ and $o$ ($x$ and $o$ are not independent, since the observation that one is likely to make will typically depend in part on the value of $x$). 

The set $M$ of possible messages is the following:
 \emph{Between 0 and 8}, \emph{between 1 and 7}, \emph{between 2 and 6}, \emph{between 3 and 5}, \emph{exactly 4}, \emph{Around 4}.\footnote{{This is of course a gross oversimplification, since we only consider messages which are `centered' on 4. The only reason we do this is that this limitation makes the model reasonably tractable and intelligible. Given that the set of \emph{observations} we consider in section \ref{sec:obs} will result in posterior distributions which are themselves centered on 4, it is likely that other ``between''-statements  that would not be centered on 4 would be generally suboptimal for the speaker compared to the messages that we include in the model.}}

 The literal listener $L^0$ is characterized by the following update rule, which defines $L^0(x=k, o=o_i|m)$, the distribution over $\langle x,o \rangle $ which characterizes the level-0 listener after she has processed the message $m$:%
\footnote{Regarding \ref{litb}, note that in contrast with Equation \eqref{eq:around2}, we use here the joint distribution $P(x,o)$ instead of just $P(x)$. We obtain this formula in the same way as we obtained the one in Equation \eqref{eq:around2}.
	Bayes'rule gives us (given the $y$ is probabilistically independent of $x$ and $o$):
	\begin{equation*}
	L^0(x=k, o=o_j, y = i \mid d(x,4) \leq y) \propto P(d(x,4) \leq y \mid x=k, o=o_j, y=i) \times P(x=k, o=o_j) \times P(y=i)
	\end{equation*}	
	
	Now, obviously, $P(d(x,4) \leq y \mid x=k, o=o_j, y=i)$ is either 0 or 1, depending on whether $d(k,4) \leq y$, and the value of $o$ does not play any role (that is, the literal meaning of the message does not carry any direct information about $o$, the observation that the speaker made, but only about $x$). So the above equation simplifies to:
	
		\begin{equation*}
	L^0(x=k, o=o_j, y = i \mid d(x,4) \leq y) \propto P(d(x,4) \leq y \mid x=k, y=i) \times P(x=k, o=o_j) \times P(y=i)
	\end{equation*}	
	
	and the rest of the computation proceeds in the same way as in Equation \eqref{eq:around2}.\\ Note that we have:

	 $L^0(x=k|m) \propto \sum\limits_{o_h \in O} L^0(x=k, o = o_h|m)\\ = \sum\limits_{o_h \in O} [P(x=k, o=o_h)\times \sum\limits_{i \geq |n-k|} P(y=i)]\\ = \sum\limits_{i \geq |n-k|} P(y=i) \times \sum\limits_{o_h \in O} P(x=k, o=o_h)\\ = \sum\limits_{i \geq |n-k|} P(y=i) \times P(x=k) = P(x=k) \times\sum\limits_{i \geq |n-k|} P(y=i).$\\ 
	 
	 Hence we recover Equation \eqref{eq:around2}.}

\ex.
\a. If $m$ is of the form \emph{between a and b} (treating ``exactly 4'' as equivalent to ``between 4 and 4''), then:
\\ $L^0(x = k ,o = o_j|m) = 0$ if $k$ is not in the interval $[a,b]$;\\ otherwise, $L^0(x = k ,o = o_j|m) \propto P(x=k, o=o_j)$.\footnote{$L^0$, when processing a ``between''-message, updates her distribution by assigning a probability 0 to values that are incompatible with the literal meaning of the message, and multiplying the probabilities of the remaining values by a constant term so that they sum up to 1.}
\b. If $m$ is the message `around 4' then:\\
 $L^0(x=k, o=o_j|m) \propto P(x=k, o=o_j)\times \sum\limits_{i \geq |4-k|} P(y=i)$ \label{litb}

\subsection{The level-1 pragmatic speaker}

The level-1 pragmatic speaker believes she talks to $L^0$. If she made the observation $o_j$, she wants to use a message $m$ such that the posterior distribution of $L^0$ over $x$ after processing message $m$ is maximally close to her own epistemic state, namely $P_{o_j}$. Now, this means she does not care about the listeners's beliefs about $o$, but only about the listener's beliefs about $x$. Let us note $L^0_m$ the distribution over $x$ of the literal listener after she has processed $m$ ($L^0_m$ is not a joint distribution over $x$ and $o$, it is a distribution over $x$ alone that results from marginalizing the conditional distribution $L^0(x=\ldots, o=\ldots \mid m)$ over $o$): $L^0_m(x=k) = L^0(x=k \mid m) = \sum\limits_{o_h \in O} L^0(x=k, o=o_h|m)$.

Based on the discussion in Section \ref{sec:speaker},  the utility function $U^1$ of the level-1 pragmatic speaker is given by:

\ex. $U^1(m,o_j) =   -D_{KL}(P_{o_j}||L^0_m)$

Finally, as is standard in RSA models, we do not assume a fully rational speaker. Rather, the speaker is only approximately rational.
That is, the higher the utility of a message, the more likely the speaker is to use it, but the best message is not used with probability 1. \PE{This is captured by means of the so-called SoftMax function. Formally, given $\vec{x}$ a sequence of real numbers $(x_1, \ldots, x_i, \ldots)$, $\textrm{SoftMax}(x_k, \vec{x}, \lambda) = \frac{\exp(\lambda \times x_k)}{\sum\limits_{x_i \in \vec{x}} \exp(\lambda \times x_i)}$. The SoftMax function maps the numbers in the list to a probability distribution, parametrized by a `\PE{rationality}'-parameter $\lambda$, a positive real number.} 
The higher $\lambda$ is, the more rational the speaker is, meaning that the probability of using the best message approaches 1 as $\lambda$ increases. The speaker $S^1$ is defined by the conditional probability of using a message out of the set $M$ of possible messages, given that a certain observation was made, and the SoftMax function is used to define this probability as follows:\footnote{
\PE{$\frac{1}{\lambda}$ is often called the \emph{temperature} parameter}. When $\lambda$ tends to infinity, this quantity tends to 1 if $m$ is the message that receives the highest utility, \PE{
as with the ArgMax function. The use of SoftMax allows a nonoptimal solution at a certain recursion level to become optimal at a later level in the recursive model presented here (cf. footnote \ref{ampl}), whereas using ArgMax would preclude this possibility. } \label{appratio}}

\ex. $S^1(m \mid o_j) = \dfrac{\exp(\lambda \times U^1(m,o_j))}{\sum\limits_{m_i \in M} \exp(\lambda \times U^1(m_i,o_j))}$


\subsection{Higher-order listeners and speakers}

Of special interest to us is the first pragmatic listener $L^1$, who interprets messages under the assumption that the author of the message is $S^1$. $L^1$ simply applies Bayes rule, and makes a \emph{joint inference} about both $x$ and $o$. Note that at this point no further inference takes place about $y$ (which enters into the interpretation of ``around'' for $L^0$). The listener $L^1$ uses the same prior distribution $P$ over $x$ and $o$ as $L^0$ (this distribution basically characterizes what was common knowledge between speaker and addressee before the speaker made any observation).

Bayes' rule gives us:

\begin{equation*}
L^1(x=k, o=o_j\mid m) \propto P(x=k, o=o_j) \times S^1(m \mid x = k, o = o_j)
\end{equation*}

Now, note that the speaker's behavior only depends on her observation --- it depends on the value of $x$ only to the extent that the observations she is likely to make are not the same across different values of $x$. The speaker does not directly observe the value of $x$, but only receives information through the observation she made, and she decides which message to use on the basis of this observation (not on the value of $x$, which she typically does not know ---  all the knowledge she has about $x$ is contained in her distribution $P_{o=o_j}$). This means that $S^1(m \mid x = k, o = o_j) = S^1(m \mid o=o_j)$. So we have:

\begin{equation*}
L^1(x=k, o=o_j\mid m) \propto P(x=k, o=o_j) \times S^1(m \mid o=o_j)
\end{equation*}

We notate $L^1_m$ the probability distribution over $x$ for the listener $L^1$ of $m$ ($L^1_m$ is the marginal distribution over $x$ for a listener who receives message $m$, not a joint distribution over $x$ and $o$). We have:

\ex. $L^1_m(x=k) = L^1(x=k \mid m) = \sum\limits_{o_h \in O} L^1(x=k, o=o_h \mid m)\\
\propto \sum\limits_{o_h \in O} P(x=k, o=o_h) \times S^1(m \mid o = o_h)$

We can then generalize this logic and define higher-level speakers and listeners as follows:

\ex. For $n \geq 1$, 
\a. $U^{n+1}(m,o_j) =-D_{KL}(P_{o_j}||L^n_m)$
\b. $S^{n+1}(m \mid o_j) \propto \exp(\lambda \times U^{n+1}(m, o_j))$

\ex. For $n \geq 1$, $L^n_m(x=k) \propto \sum\limits_{o_h \in O} P(x=k, o=o_h) \times S^n(m \mid o_h)$

As we shall see in the next section, the effect of ``around'' tends to be magnified for higher-level listeners and speakers  compared to what happens at the level of the literal listener, in that the bias towards central values is increased.

\section{A concrete implementation of the interactive model} \label{concrete}

We now provide a concrete implementation of the model we have just described.\footnote{\bs{
The code for this implementation ({\sf R} scripts) as well as that of the alternative
models discussed in section 8 can be downloaded from
\url{https://github.com/ BenSpec/ScriptsAround.}}} 
The goal here is not to propose a realistic model --- as we shall see, some choices will be quite arbitrary --- but to illustrate the point that such an explicit model predicts that the basic effect we have been discussing (the fact that  ``around''-statements can be used to convey the shape of a probability distribution) can be amplified by pragmatic reasoning.

\subsection{Observations and prior probability distributions} \label{sec:obs}

As we saw, we assume that the variable of interest $x$ ranges over the natural numbers between 0 and 8. The model includes nine observations. From the point of view of the model, the only thing that matters is the effect of an observation $o_j$ on an observer who starts with a prior probability distribution $P$ over $x$, i.e. we need to specify $P_{o_j}$, i.e. the different epistemic states the speaker could be in after having made an observation. Here we are interested in comparing the speaker's behavior in epistemic states that have the same support but a different  `shape' (from uniform distributions to distributions biased towards central values). The nine observations we consider are specified in Table \ref{obs}, where each column is the probability distribution induced by a specific observation, and each line is a possible value for $x$.\footnote{Values are rounded to 2 digits (here and elsewhere), but actually sum up to 1 in all columns (the value $0$ thus sometimes corresponds to a non-zero but extremely small value). The choice of these nine distributions is for the sake of illustration, and we could stipulate other values in the case of peaked distributions.}

%
\begin{table}[H]
\caption{Posterior distributions resulting from observations ($P(x|o$))}\medskip \label{obs} \vspace{-.5cm}
\[
\includegraphics[scale=.65]{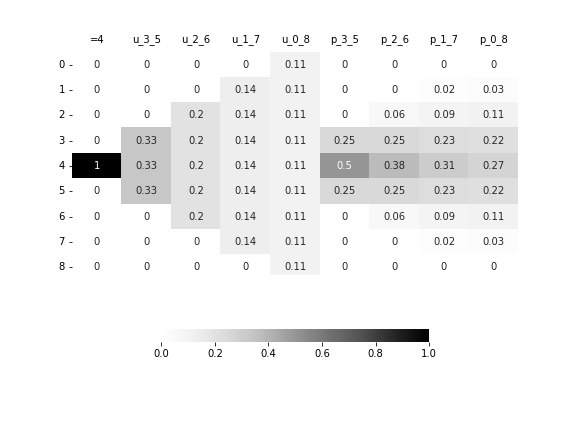} \vspace{-2cm}
\]
\end{table}

The first five columns correspond to five observations that give rise to a uniform distribution on an interval centered on 4 (from the observation $=4$, where the speaker can conclude from her observation that $x=4$, to the observation $u\_0\_8$, which results in a uniform distribution on the full range of $x$). The last four observations correspond to distributions with supports $[3,5]$, $[2,6]$, $[1,7]$ and $[0,8]$, respectively, with a bias towards values closer to 4. These were obtained from binomial distributions with parameter 0.5, shifted to the relevant intervals. Thus, an observation named $\mathsf{u\_a\_b}$ corresponds to a posterior distribution which is uniform on its support $[a, b]$, while $\mathsf{p\_a\_b}$ corresponds to a posterior distribution with support $[a,b]$ which is `peaked', in the sense of being biased in favor of more central values.

To build the joint distribution on $\langle x,o\rangle $, we will assign first probabilities to each observation, and derive the joint distribution from the fact that

 \begin{equation*}
 P(x=k,o=o_i)=P_{o_i}(x=k) \times P(o_i). \label{Joint}
 \end{equation*}
 
  The resulting marginal probability distribution over $x$ will then be given by 
  
  \begin{equation*}P(x=k) = \sum\limits_{o_i \in O} P(o_i)\times P_{o_i}(x=k). \label{marg} \end{equation*}

Now, values close to $4$ belong to the support of more distributions than values further from $4$, and all the distributions induced by an observation are either uniform or biased towards values closer to $4$. As a result of this setup, the marginal distribution over $x$ will itself be non-uniform, and biased towards central values.\footnote{This is by no means a necessary choice. But to make an already complex model reasonably tractable, we chose to restrict the set of possible observations to those that are `centered' on 4, as this suffices to make our main point. \bs{Importantly, we can build a richer model, which includes precise observations for each possible value for $x$, with the result that the prior distribution on $x$ is near uniform (or even such that the prior on the central value is less than the one on more peripheric values). Simulations show that this does not make any qualitative difference to the model's predictions. The code for this richer model is available at \url{https://github.com/BenSpec/ScriptsAround}.}} We chose to assign higher probabilities to observations that yield distributions with a larger support, so as to not penalize too much peripheral values. Specifically, we assigned the following weights to each observation, which we then normalized to get a probability distribution:

\begin{table}[H]\centering
\caption{Probabilities over observations ($P(o=o_i)$).} \label{table:probobs}\medskip
 \resizebox{\textwidth}{!}{
\begin{tabular}{r|ccccccccc}
	\toprule
Observation	& $\mathsf{=4}$ & $\mathsf{u\_3\_5}$ & $\mathsf{u\_2\_6}$ & $\mathsf{u\_1\_7}$ & $\mathsf{u\_0\_8}$ & $\mathsf{p\_3\_5}$ & $\mathsf{p\_2\_6}$ & $\mathsf{p\_1\_7}$ & $\mathsf{p\_0\_8}$ \\ 
	\midrule
Non-normalized Weight & 1 & 4 & 16 & 64 & 256 & 1 & 4 & 16 & 64 \\ 
Probability & $\frac{1}{426}$ & $\frac{2}{213}$ & $\frac{8}{213}$ & $\frac{32}{213}$ & $\frac{128}{213}$ &  $\frac{1}{426}$ & $\frac{2}{213}$ & $\frac{8}{213}$ & $\frac{32}{213}$ \\ 
	\bottomrule
\end{tabular}}
\end{table}

The resulting marginal probability distribution on $x$ (given by the preceding equation) is the following:

\begin{table}[H]
	\centering
	\caption{Prior Probability Distribution over $x$.} \label{n}\medskip
	\begin{tabular}{c|ccccccccc}
		\toprule
	$x$	& 0 & 1& 2 & 3 & 4 & 5 & 6 & 7 & 8 \\ 
		\midrule
	
		$P(x)$ & 0.07 & 0.09 & 0.12 & 0.14 & 0.16 & 0.14 & 0.12 & 0.09 & 0.07 \\ 
		\bottomrule
	\end{tabular}
\end{table}

\subsection{Other parameter of the models}

The variable $y$, which determines the size of the interval denoted by ``around 4'' and enters into the interpretation of ``around 4'' at the level of the literal listener, ranges from 0 to 4 (recall that $y = i$ means that the intended interval for ``around 4'' is [4-i, 4+i]). We assume that the prior probability distribution over $y$, which is used by the literal listener, is uniform.

As discussed in section \ref{setup}, we are assuming that the speaker has six messages at her disposal (\emph{Exactly 4}, \emph{between 3 and 5}, \emph{betwen 2 and 6}, \emph{between 1 and 7}, \emph{between 0 and 8}, \emph{around 4}). We set the `\PE{rationality} parameter' entering into the utility function of the speaker to 10. \PE{This choice is again stipulated for the sake of illustration, and other values could be selected}.

\subsection{Predictions of the model}

First we look at the literal listener $L^0$, in Table \ref{L0}. Each cell of the table represents the probability assigned by $L^0$ to a number $0$ and $8$ after having interpreted a given message (``b. 3 and 5'' represents ``between 3 and 5'').

\begin{table}[H]
\caption{\footnotesize{Probabilities assigned by $L^0$ to each value for $x$ after processing a message $m$ ($L^0(x\mid message$))}} \vspace{-0.6cm} \label{L0}\medskip
\[
\includegraphics[scale=.65]{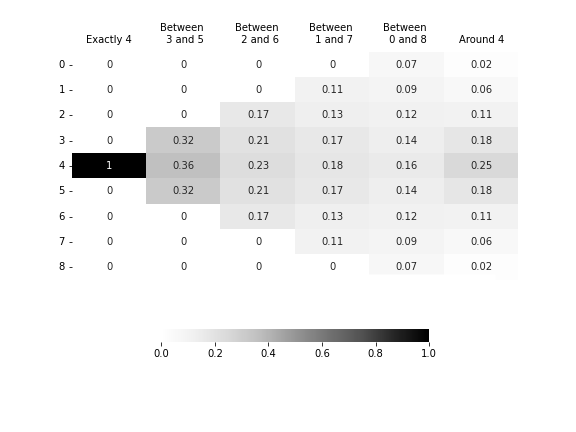} \vspace{-2cm}
\]
\end{table}


Note that the message \emph{between 0 and 8} is in fact completely uninformative. As expected, then, the posterior distribution that results from this message is identical to the prior distribution (cf. Table \ref{n}) . All other ``between''-messages assign 0 to the numbers they exclude. Finally, the ``around''-message does not assign $0$ to any number, but results in a distribution that is much more biased in favour of central values than the prior distribution was, as expected (one can compare it with the column for \emph{between 0 and 8}, which corresponds, as noted, to the prior distribution). Note also that it assigns a higher value to $4$ than the posterior distribution obtained after processing \emph{between 2 and 6}, \emph{between 1 and 7}, and \emph{between 0 and 8}, following the logic discussed in section~\ref{sec:ratio}. So it might end up being the best message to use for a speaker who is uncertain but assigns a high probability to central values. {In Fig.~\ref{L0}, we represent $L^0$'s distribution after processing the messages \emph{between 1 and 7} and \emph{around 4}, together with the distribution resulting from the observation $ p\_1\_7$, i.e. a peaked distribution with support $[1, 7]$  -- the distribution induced by \emph{around 4} (in green in Fig. \ref{L0}) is intuitively closer to the speaker's distribution (in blue) than the one induced by \emph{between 1 and 7} (in orange), despite having a broader support.
}

\begin{figure}[H]
\caption{\footnotesize{$L^0$'s distributions after `b. 1 and 7' and `around', compared with distribution induced by obs. $p\_1\_7$}} \vspace{-0.6cm} \label{L0}\medskip
\[
\includegraphics[scale=.62]{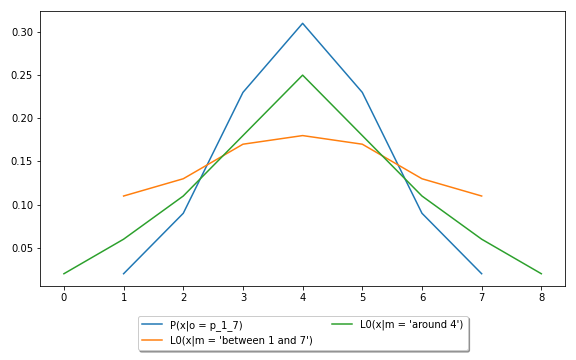} 
\]\vspace{-0.3cm}
\end{figure}

We now turn to the behavior of the level-1 speaker, who talks to this listener. The following table represents the probability of using a certain message depending on the observation that the speaker made.

\begin{table}[H]
\caption{\footnotesize{$S^1$'s probability of choosing a message depending on the observation made ($S^1(m \mid o)$).}} \label{S1}\medskip
 \resizebox{\textwidth}{!}{
\begin{tabular}{c|cccccc}
  \toprule
\backslashbox{{\footnotesize Observation}}{\footnotesize Message} & Exactly 4 & b. 3 and 5 & b. 2 and 6 & b. 1 and 7 & b. 0 and 8 & around 4 \\ 
  \midrule
=4 & 1.00 & 0.00 & 0.00 & 0.00 & 0.00 & 0.00 \\ 
  u\_3\_5 & 0.00 & 0.98 & 0.01 & 0.00 & 0.00 & 0.01 \\ 
  u\_2\_6 & 0.00 & 0.00 & 0.82 & 0.07 & 0.02 & 0.09 \\ 
  u\_1\_7 & 0.00 & 0.00 & 0.00 & 0.69 & 0.16 & 0.15 \\ 
  u\_0\_8 & 0.00 & 0.00 & 0.00 & 0.00 & 0.93 & 0.07 \\ 
  p\_3\_5 & 0.00 & 0.97 & 0.01 & 0.00 & 0.00 & 0.01 \\ 
  p\_2\_6 & 0.00 & 0.00 & 0.68 & 0.06 & 0.01 & 0.25 \\ 
  p\_1\_7 & 0.00 & 0.00 & 0.00 & 0.27 & 0.06 & \textcolor{red}{\textbf{0.66}} \\ 
  p\_0\_8 & 0.00 & 0.00 & 0.00 & 0.00 & 0.14 & \textcolor{red}{\textbf{0.86}} \\ 
   \bottomrule
\end{tabular}}
\end{table}

When the speaker's distribution is uniform across a certain interval (first five lines), the speaker has a very high probability of using the corresponding \emph{between}-statement. 
If the speaker's distribution has support $[0,8]$ but is biased towards central values (last line), the speaker prefers the \emph{around}-statement. Importantly, even when the speaker is able to categorically exclude 0 and 8 and has a distribution which is biased in favor of central values ($\mathsf{p\_1\_7}$), she prefers to use the \emph{around}-statement rather than the corresponding \emph{between}-statement, following the logic of what we discussed in Section \ref{speaker choice}.  She has also a non-insignificant probability of using the \emph{around}-statement when her distribution is a peaked distribution with support is $[2,6]$

Now, we consider the 1st-level pragmatic listener (Table \ref{L1}). This listener knows that the speaker's choice of message is governed by Table \ref{S1}. So when hearing the \emph{around}-statement, she will infer that the speaker is most likely in the epistemic state that results from the observations $\mathsf{p\_0\_8}$ or  $\mathsf{p\_1\_7}$.  She will update her distribution over $x$ on this basis. \vspace{-.3cm}

\begin{table}[H]
\caption{\footnotesize{Probabilities assigned by $L^1$ to each value for $x$ after processing a message $m$ ($L^1(x\mid message$))}}\vspace{-0.6cm} \medskip
 \label{L1}
\[
\includegraphics[scale=.65]{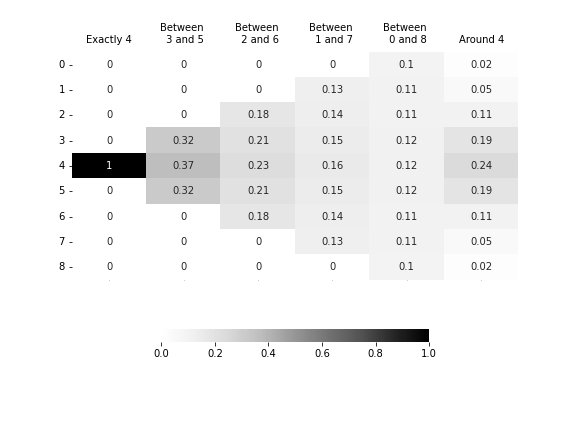} 
\] \vspace{-2cm}
\end{table}

While the posterior distribution of $L^1$  (cf. Table \ref{L1}) after processing \emph{around 4} is slightly less peaked than it was for $L^0$, the posterior distributions induced by the messages \emph{between 0 and 8}, \emph{between 1 and 7} and \emph{between 2 and 6}  are themselves significantly flatter (more uniform) than they were for $L^0$, so the contrast in interpretation between statements based on \emph{between} and the one based on \emph{around} is maintained (and even amplified if we compare the ratios, across distributions, between central values and peripheral values that have a non-null probability).

 We can then look at even higher-order listeners and speakers. After a few iterations, we reach a near-steady state where further iterations do not change anything significantly.  {Table \ref{table-L4} shows $L^4$'s posterior distribution over $x$ after each message.}
 
\begin{table}[H]
\caption{\footnotesize{Probabilities assigned by $L^4$ to each value for $x$ after processing a message $m$ ($L^4(x\mid message$))}}\vspace{-0.6cm} \medskip
 \label{table-L4}
\[
\includegraphics[scale=.65]{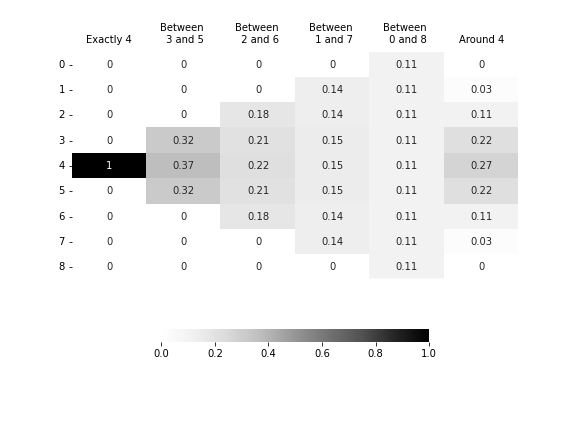} 
\] \vspace{-2.7cm}
\end{table}

\noindent We can see that the basic effect we already saw at lower levels of recursion ($L^0$ and $L^1$) is now stronger, in that the distribution induced by \emph{around 4} is more peaked for $L^4$ than for $L_0$ and $L_1$. {For this reason, the level-5 speaker, who believes she is talking to $L^4$, will now prefer the message \emph{around 4} over \emph{between 2 and 6} when the support of her distribution is $[2,6]$ but is peaked (corresponding to the observation $\mathsf{p\_2\_6}$). This is because, as illustrated in Fig.~\ref{L4}, $L^4$'s distribution over $x$ after processing \emph{around 4} (green curve in Fig.~\ref{L4})~is a better match to the distribution induced by observation $\mathsf{p\_2\_6}$ (blue curve), 
 despite the fact that its support includes numbers outside of $[2,6]$.}
%
%
%
%

 \begin{figure}[H]
\caption{\footnotesize{$L^4$'s distributions after `b. 2 and 6' and 'around', compared with distribution induced by obs. $p\_2\_6$}} \vspace{-0.6cm} \label{L4}\medskip
\[
\includegraphics[scale=.62]{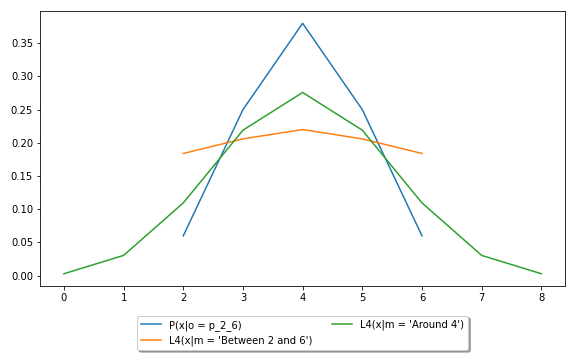} 
\]
\end{figure}

\noindent Table \ref{S5} displays the behavior of the level-5 speaker, who believes she is talking to the level-4 listener. 
  
\begin{table}[ht]
\centering
\caption{\footnotesize{$S^5$'s probability of choosing a message depending on the observation made ($S^5(m \mid o)$)}} \label{S5}\medskip
\resizebox{\textwidth}{!}{\begin{tabular}{c|cccccc}
  \toprule
\backslashbox{\footnotesize Observation}{\footnotesize Message} & Exactly 4 & b. 3 and 5 & b. 2 and 6 & b. 1 and 7 & b. 0 and 8 & around 4 \\ 
  \midrule
=4 & 1.00 & 0.00 & 0.00 & 0.00 & 0.00 & 0.00 \\ 
  u\_3\_5 & 0.00 & 0.96 & 0.01 & 0.00 & 0.00 & 0.03 \\ 
  u\_2\_6 & 0.00 & 0.00 & 0.78 & 0.03 & 0.00 & 0.19 \\ 
  u\_1\_7 & 0.00 & 0.00 & 0.00 & 0.87 & 0.08 & 0.06 \\ 
  u\_0\_8 & 0.00 & 0.00 & 0.00 & 0.00 & 1.00 & 0.00 \\ 
  p\_3\_5 & 0.00 & 0.96 & 0.01 & 0.00 & 0.00 & 0.03 \\ 
  p\_2\_6 & 0.00 & 0.00 & 0.41 & 0.01 & 0.00 & \textcolor{red}{\textbf{0.57}} \\ 
  p\_1\_7 & 0.00 & 0.00 & 0.00 & 0.05 & 0.00 & \textcolor{red}{\textbf{0.94}} \\ 
  p\_0\_8 & 0.00 & 0.00 & 0.00 & 0.00 & 0.01 & \textcolor{red}{\textbf{0.99}} \\ 
   \bottomrule
\end{tabular}}
\end{table}

%
%

We see that the speaker will prefer the \emph{around}-message when her observation leads to a peaked distribution on the intervals $[2,6]$, $[1, 7]$ and $[0,8]$, and that when her distribution is uniform, she goes for the \emph{between}-message that corresponds to the support of her distribution. 
The recursive aspect of the model led to an amplification of the basic phenomenon observed at the level of the literal Listener $L^0$ and the first-level pragmatic speaker $S^1$. In particular, the level-5 speaker now uses the ``around''-statement also when her distribution is peaked with support $[2, 6]$.\footnote{{As noted in footnote \ref{appratio}, modeling the speaker as not being fully rational is necessary in order to derive this effect: with a fully rational speaker, the probability that $S_1$ picks the \emph{around}-message to communicate a peaked distribution with support $[2,6]$ would be 0, making the message not interpretable by $L_1$ and subsequent listeners.} \label{ampl}}

\section{Comparison with the Lexical Uncertainty approach}\label{sec:comparison}

Our full model is couched in the RSA framework for pragmatics. {Within the RSA framework, one of the most influential approaches to vagueness is \posscite{lassiter2017adjectival} proposal for gradable adjectives. The closely related Lexical Uncertainty model of \citealt{bergen2016pragmatic} (LU for short) provides a general model of meaning underspecification.}
However, this literature does not discuss one of the main points of our paper: the fact that vague language might allow a speaker who is not fully informed about some topic under discussion to communicate information about the \emph{shape} of her probability distribution. As it turns out, we can show that without fundamental modifications, the LU model, \bs{as spelled out in \citet{bergen2016pragmatic}}, is unable to reproduce the qualitative predictions of our model, especially regarding the speaker's behavior.

Concerning speaker uncertainty, \citet{lassiter2017adjectival} simply do not incorporate in their model the possibility that the speaker is not fully informed about the value of interest (say, someone's height, in relation to the use of \emph{tall}). They consider a speaker who knows, say, Mary's height, and can use messages such as \emph{Mary is tall}, \emph{Mary is not tall}, \emph{Mary is short}, \emph{Mary is not short}. The goal of the model is to predict the \emph{interpretation} of such messages on the listener's side, and the listener is assumed to reason under the assumption that the speaker is fully informed. Their paper focuses on how, when processing such a message, the first-{level} pragmatic listener {$L^1$} updates their probability distribution over Mary's height. 

\bs{Even though this is not done in \citet{lassiter2017adjectival}, 
\posscite{bergen2016pragmatic} LU model, which can  be viewed as an extension of \citet{lassiter2017adjectival}, deals with non-fully informed speakers}.\footnote{It was in part developed to deal with so-called Hurford Disjunctions, i.e. sentences like \emph{Mary ate some or all of the cookies}, which imply that the speaker is not fully informed.} It is straightforward to apply the LU model to ``around'' in the general case where the speaker is not fully informed, by treating the size of the intended interval in the same way as the threshold for gradable adjectives is treated in LU models. {Since such a model defines the utility function of the speaker in terms of Kullback-Leibler divergence,}\footnote{{Technically speaking, many RSA models are written without explicit reference to Kullback-Leibler divergence, but it is easy to show, that they are fully equivalent to models that use the Kullback-Leibler divergence in the utility function  of the speaker.}}{~one might think --- and this was our initial expectation --- that it would reproduce the qualitative predictions of our model, especially the prediction that the probability of using an``around''-sentence might depend on the \emph{shape} of the speaker's distribution, rather than just its support}.  This is in fact not the case,  for a fundamental reason, namely the following mathematical fact, proven in Appendix \ref{app:LU}:%

\ex. Let two observations $o_1$ and $o_2$ be such that the supports of the
conditional distributions $P(x=k|o_1)$ and $P(x=k|o_2)$ are identical (that is
$P(x=k|o_1) > 0$ iff $P(x=k|o_2) > 0$). In the LU model, for every message $m$, at every step of the recursion, we have $S(m|o_1) = S(m|o_2)$. \label{support}

This means that the speaker's choice of a message only depends on the \emph{support} of her distribution, not on its shape, and so one core idea of our own proposal cannot be captured in the LU model. This is not to say that the LU model predicts none of the effects we discuss. The first-level listener, in the LU model, does end up with a posterior distribution that favors central values after processing an ``around''-statement (though, in our simulation, to a much lower extent than in our model). However, the speaker does not take this fact into account when choosing her message, and as a result this effect on the listener tends to {fade away} when we move higher {up} in the recursive sequence of listeners (because in contrast with our model, pragmatic listeners can only draw inferences about the support of the speaker's distribution, not its shape).

To be more precise, the LU model with a non-fully informed speaker differs from our own in two main respects.

First, semantic underspecification (in our case, the size of the interval for \emph{around}) is dealt with differently. In the LU model, applied to \emph{around}, the literal listener {$L^0$} is relativized to a specific interpretation function, and would interpret an ``around $n$''-statement as meaning ``between $a$ and $b$'', where $a$ and $b$ are set by the interpretation function. So there are as many literal listeners as there are interpretation functions (i.e. ways of interpreting ``around'', since the interpretation function does not matter for other messages). Likewise, the first-level speaker is relativized to an interpretation function, and chooses her message on the assumption that the listener she is talking to is a literal listener relativized to the same interpretation function. So at the level of the literal listener and the first-level speaker, nothing interesting happens for an ``around''-sentence, which is just treated in the same way as  a ``between''-statement. It is at the level of the first-{level} pragmatic listener (and similarly for speaker) that reasoning about the interpretation function (i.e. the intended interval for \emph{around}) takes place, in the sense that this first-level pragmatic listener is no longer relativized to a specific interpretation function, but reasons probabilistically about the interpretation function, treated as a random variable. {In our model, the listener of level 0 is right away interpreting ``around'' by taking into account its multiplicity of more precise interpretations.}

The second difference is that in the LU model, the utility function of a non fully informed speaker is subtly different from ours, \bs{and also from a number of other RSA models, including \cite{goodman2013knowledge}} In the LU model, the utility function is defined in terms of the Kullback-Leibler divergence  of the listener's joint distribution over $\langle x,o\rangle $ from the speaker's joint distribution over $\langle x,o\rangle$. 
In contrast, our speaker, \bs{just like in the initial model proposed in \cite{goodman2013knowledge}}, wants to minimize the Kullback-Leibler divergence of the listener's marginal distribution over $x$ from the speaker's marginal distribution over $x$ \bs{(exactly as discussed in, e.g., \citealt{ProbLang}, Chapter 2 and Appendix II)}.\footnote{\bs{It seems that this subtle difference between the utility function defined in \cite{goodman2013knowledge} and the one used in \cite{bergen2016pragmatic} has not been discussed at all in previous literature. To complicate the matter, as pointed out to us by Michael Franke and as hinted in \cite[Chapter 2]{ProbLang}, in their actual implementation, \cite{goodman2013knowledge} do not seem to have used the notion of utility they define in their paper, but to actually implement yet a different model. Thanks to Michael Franke for very helpful discussions.}}

Hence, the utility function used in the LU model views the speaker as caring not only about bringing the listener's distribution over the variable of interest as close as possible to hers, but also about communicating to the listener her private epistemic state about this variable. In contrast, the utility function in our model views the speaker as wanting to bring the listener's distribution over the variable of interest as close to hers, but not as caring about whether the listener correctly identified her epistemic state. \bs{Consider for instance a situation where the speaker believes that either $A$ or $B$ is true (where $A$ and $B$ are mutally incompatible), and assigns to each a 50\% probability. On the basis of the utility function of the LU model, the goal of the speaker is to make the listener assign a 50\% probability to each of ${A, B}$, and, on top of that, to ensure that the listener knows that the speaker herself assigns these probabilities. On the basis of the utility function used in our model, unless the speaker's beliefs are explicitly under discussion, the speaker's goal is \textit{only} to make the listener assign at 50\% probability to each of $A$ and $B$, and the speaker \textit{does not mind} if the listener, for instance, wrongly believes that the speaker is fully knowledgeable.}

{In many RSA models, the choice between these two options does not greatly affect qualitative predictions. And indeed, from the standpoint of our model, simulations show that if we use the utility function of the LU model, we can still reproduce the qualitative predictions of our model ({cf. model described in Appendix \ref{sec:variant_our}}). From the standpoint of the LU model, however, it turns out that this choice is highly consequential. One can construct ({cf. Appendix \ref{sec:variantLU}}) a version of the LU model where the utility function is defined as in our model in terms of the KL-divergence of the listener's and speaker's distributions over $x$ (rather than their joint distributions over $\langle x,o\rangle$), but in which the result in \ref{support} no longer holds: simulations show that, at least for very high values of the \PE{rationality} parameter, the amended model can make predictions which are qualitatively similar to ours.} {We should note that this modified version of the LU model, just like the standard LU model, reduces to \posscite{lassiter2017adjectival} model if we only consider a fully-informed speaker. While \citet{lassiter2017adjectival}, as noted, does not consider a non-fully informed speaker, the LU model is just one possible way of generalizing it, and the alternative version of the LU model with non-fully informed speaker described in Appendix  \ref{sec:variantLU} is another one, \bs{which is by itself fully consistent with the rest of the RSA literature.}}


The upshot of this discussion is that it is the combination of a specific architectural choice (postponing to $L^1$ the listener's reasoning about the size of the interval for ``around'') and the choice of a specific utility function that makes the LU model unable to predict that the shape of the speaker's distribution, and not just its support, plays a role in the speaker's choice of a message. To forestall the limitation we state in \ref{support}, one can either drop the architectural feature (e.g., by moving to a model like ours where the literal listener already treats the interval size for \emph{around} as a random variable), or change the utility function along the lines of our model (and in line with a number of other RSA models). Our discussion thus provides an argument for potential amendments to \bs{one} standard treatment of semantic underspecification in RSA models.\footnote{{Recently, another RSA model with meaning underspecification has been proposed by Franke and Bergen (\citealt{franke2020theory}), the \emph{Lexical Intention} model (LI for short). In this model, the speaker  chooses simultaneously a message and a meaning for the message, while in the LU model $S^1$ is relativized to a certain interpretation function but does not \emph{choose} one. The model has been applied only to cases where the speaker is assumed to be fully informed. If we extend it on the basis of the  utility function used in the LU model for the more general case of a non-fully informed speaker, the utility (at level 1) of a pair $<$message, meaning$>$  is defined by exactly the same formula as in the LU model, even though it has a different interpretation. This ensures that the limitation result  proved in Appendix {\ref{app:LU}}  for the LU model also holds for this version of the LI model. In fact, in this extended version of the LI model, within the setup described in section \ref{concrete}, the `around $4$' statement is entirely uninformative, i.e. the pragmatic listener does not gain any information from it. \bs{However, if in fact we use a utility function identical to the one described in Appendix  \ref{sec:variantLU}, the resulting model makes predictions that are qualitatively similar to ours, at least for very high values for $\lambda$. Simulations can be found at \url{https://github.com/BenSpec/ScriptsAround}}}}

\section{Limitations}\label{sec:further}

\subsection{Rounding and Granularity} \label{sec:rounding}
While our model predicts the contrasts highlighted in section \ref{sec:motivations} regarding the use of ``around'' and ``between'', the central assumption we made of a speaker who is uncertain about the state of the world sets aside further facts concerning the use of ``around''. 

The first of those concerns rounding, namely cases in which the speaker is perfectly informed about the numerical value of interest, but may nevertheless choose to use ``around'' instead of reporting the exact value. Consider a teacher who knows that 19 children attended her class. When asked ``how many children did you have in class today?'' she may respond by uttering: ``around 20 children''.\footnote{``Approximately 20'' may be more natural to use than ``around 20'' in case the speaker is perfectly informed and does rule out 20; however, we think it is possible to use ``around'' in the sense of ``approximately''. There are obviously subtle differences in meaning between``around'', ``about'', and ``approximately'', which we set aside in this paper.}
In that case, it would be incorrect for the listener to infer that {the speaker considers 20 to be the most likely value, since by assumption the most likely value is 19}. 
\bs{Furthermore, if the speaker is believed to be well informed, ``around 20'' suggests that the true value is \emph{not}~20, since otherwise the speaker would have simply said ``20''.  
This appears to contradict our model.
\footnote{\PE{It might even be common knowledge that the target number for ``around'' is ruled out. An example from an anonymous referee is: ``there might be a rule that congregations of a cult must gather in odd numbers. One member can brag about their congregation, saying: We have typically around 100 members".}}

However, these kinds of uses always seem to involve round numbers. Thus, in the very same context, it would {be} extremely strange for the speaker to respond with ``around 18 children''. For the latter, the listener 
{will} make the inference that the speaker is not perfectly informed about the state of the world, because {18}  \emph{cannot be} a round number {when the finest possible granularity scale is that of natural numbers} (contrast with ``around 18 kilometers'' to report  a distance run by bike, which may be used to round off a distance expressible with finer granularity in hundreds of meter or in meters).}\footnote{Reporting ``around 18 children'' is odd but not ruled out, for instance if the speaker tries to remember how many children attended class, by remembering how many children sat in each row, adds up the numbers, and wants to convey that the value obtained may be inaccurate.}

There is, therefore, a clear interaction between the use of ``around'' and considerations of granularity \bs{and roundness}.\footnote{It is also plausible that granularity considerations influence the interpretation of ``around''-statements even when the speaker is not assumed to be fully knowledgeable. For instance, it might happen that ``around 40'' makes the interval [30, 50] particularly salient. One way to capture such an effect in our model would be to assume that the prior probability distribution over the intervals denoted by ``around'' is influenced by granularity considerations and, instead of being uniform, gives more weight to intervals contained in $[30, 50]$, which would incorporate some of the insights from the literature (e.g., \citealt {krifka2007approximate, solt2014alternative, solt2017preference})}\label{fngran}
{So far, our} model does not involve any consideration of granularity and roundess, and excludes the possibility that a fully informed speaker will use an ``around''-sentence with a significant probability}. \bs{It is however possible in principle to enrich our model in order to include such considerations. One natural possibility would consist in adding a cost term in the utility function of the speaker, and to assign lower costs to messages involving round numbers (compatibly with \citealt{solt2017preference}'s data on the fact that round numbers are easier to remember and manipulate than non-round numbers).} \bs{This would express the idea that speakers have a preference for round numerals, so that they might prefer an ``around''-statement with a round numeral over a more informative statement involving a non-round numeral. If we add a cost term to our utility function and make round numerals less costly than non-round numerals, a fully-informed speaker might, in certain cases, prefer to use \emph{around} with a round numeral over using an unmodified non-round numeral, because the loss of informativity incurred by not stating the precise (non-round) value can be smaller than the gain obtained on the cost side. As a result, a pragmatic listener will not exclude the possibility that the speaker might be fully informed if she said, e.g., ``around 20''. Moreover, if in fact the listener antecedently believes that the speaker is fully informed, she will think that probably the true value is not 20 (since ``20'' would have been more informative than ``around 20'' if the speaker were sure that the true value is 20).  Of course, to fully address such complex interactions between roundness, informativity, and the speaker's epistemic state, we would need to build such an extended model and conduct an in-depth analysis of its detailed predictions, which we leave to another occasion}.\footnote{For a preliminary investigation of rounding in cases in which the speaker is perfectly informed, we refer to \cite{mortier2019master},
 {which develops a model within the Lexical Uncertainty framework where messages with round numbers are less costly than those using non-round numbers, so that a fully informed speaker might choose  an ``around''-message in order to avoid using a non-round number.}}


\subsection{Common priors}

Another limitation of the model concerns the common prior assumption. This limitation is not specific to our approach, it is shared by RSA approaches and {by most game-theoretical set-ups. It is needed for the recursive definition of listeners and speakers to make sense from a normative point of view.} It assumes that the interlocutors have access to their own probability distribution, but also that the speaker and the listener shares the same prior probability distribution over the variable of interest before the speaker makes a private observation about the state of the world. Those assumptions are obviously disputable, {as they are most of the time violated in real life}. Although we do not need to assume a strong form of introspection to make sense of the use of personal probabilities, the common prior assumption is much stronger. In practice, and more realistically, distinct agents may rather have priors about each other's priors, and could very well be mistaken. 

We believe that a distinct model could be designed along those lines, though it would have to be significantly more complex.\footnote{{Most works in economic theory and game-theory accept some form of the Common Prior Assumption, and dispensing with this assumption leads to non-trivial conceptual and theoretic challenges, as discussed in \citet{morris1995common}.}} For our purposes, however, we think it is sufficient to produce a worked out model of the contrast between ``around'' and ``between'' along the lines we suggested, setting aside further refinements.


\section{Semantic flexibility}\label{sec:flex}

\bs{By showing that vague language can be more informative than precise language and can thereby serve a specific communicative purpose, our approach also \pe{gives us a fresh perspective on whether vagueness is better conceived as an epistemic or as a semantic phenomenon (\citealt{sorensen1988blindspots,williamson1994vagueness, wright1995epistemic}}). 

\pe{Technically speaking}, our model is \emph{compatible} with the epistemic theory. That is, it is coherent with our model to assume that there is a fact of the matter as to the value taken by the parameter of interpretation for \emph{around} (i.e. the half-length of the interval).  If so, the prior probability over this parameter (used by the literal listener) would represent the listener's uncertainty about a determinate fact -- the `real' truth-conditions of the relevant \emph{around}-statement, in line with epistemicism (\citealt{lassiter2017adjectival} make a similar point). This, however, is not the only possible interpretation of our model, and is not in fact the most natural one. In particular, on our view, the \emph{function} of a word like \emph{around} is to introduce semantic underdeterminacy, allowing us to communicate probabilistic information. In agreement with \citealt{wright1995epistemic}, we find it highly counterintuitive to think that the core meaning of \emph{around} (and similar approximators) is in fact precise, given that its function seems to introduce vagueness. As Wright puts it \citep[153-154]{wright1995epistemic},  ``\emph{the role of such particles seems unquestionably to be to introduce some conveniently indeterminate degree of flexibility}" (our emphasis).

We believe that our model in fact vindicates Wright's core intuition. On our view, the lexical entry for \emph{around} is relativized to an open parameter, and so an \emph{around}-statement does not by itself express a proposition. Moreover, we do not need to assume that there is a `true' value for this parameter. The literal listener's interpretation only depends on a probability distribution over this parameter, and the only thing that matters for communication to be successful is that the speaker knows how the literal listener interprets her sentence. That is, in our model, the proposition expressed by an ``around''-statement relative to a certain fixed value of the parameter does not play any direct role.  This means that we could as well relativize the meaning of an ``around''-statement to a probability distribution over the parameter $y$, \bs{or even} characterize it directly in terms of an interpretation rule for the literal listener (the one expressed by Equation \ref{eq:around2}). 

In fact, we could in principle derive our rule for the literal listener without even assuming a lexical entry for \emph{around} that refers to a threshold (here the half-length of the denoted interval). We could instead use a fuzzy logic approach, where an ``around''-sentence would denote a (possibly context-dependent) function that maps every world to a number in $(0,1)$, and incorporate it within an RSA model (similarly to \posscitealt{van2021probabilistic} proposal for quantifiers, which aims to capture typicality effects). We would define a literal listener as in the standard RSA model, by $L_0(w|m) \propto P(w) \sem{m}(w)$, but $\sem{m}(w)$ could be any number from $0$ to $1$ (in contrast to  `classical' models where $\sem{m}(w)$ denotes a truth-value, i.e. $0$ or $1$).  To the extent that the gradient meaning that would be assigned to \emph{around n} would favor values closer to $n$, the ratio inequality discussed in section \ref{sec:ratio} would hold, and, keeping all other aspects of the model constant, we would make the same qualitative predictions.\footnote{{In fact, as noted in Appendix \ref{app:original}, we initially started with a different model for the literal listener, with no major change in terms of qualitative predictions.}} Furthermore, it is even possible to assign to ``around $n$'' a gradient semantics of this sort in such a way that the resulting model would be a notational variant of ours.\footnote{{Here is a way to do this:

\ex. $\sem{\text{around n}} = \lambda x. \dfrac{\sum\limits_{n \geq i \geq |n-x|} f(i)}{\sum\limits_{0 \leq x' \leq 2n} \sum\limits_{n \geq i \geq |n-x'|} f(i)}$, where $f$ is some function from $(0, n)$ to a null or positive number.

\ex. $L_0(x=k, o = o_j|m) \propto P(x=k, o=o_j)\sem{m}$

This is exactly equivalent to our official model when we identify $f(i)$ with $P(i)$.}}
In this variant, there clearly could not be any `fact of the matter' as to what is the `true' interval denoted by \emph{around n}, since the lexical entry for \emph{around} would simply not involve any interval. That being said, our proposal in terms of a free parameter has two potential advantages: the update rule for the literal listener is derived through rational, Bayesian inference, and we can keep to a simple, bivalent compositional semantics when we apply our proposal to more complex sentences (where an `around $n$' statement is embedded in a larger sentence).
 
 {In any case, however the literal listener's rule is derived,} the point we are making here is that {our treatment} of ``around'' and similar approximating expressions does indeed guarantee the ``conveniently indeterminate degree of flexibility'' claimed by Wright, in a way that no truth-conditionally precise surrogate can provide. While we agree with the epistemicist that the use of vague expressions is constrained by general maxims of knowledge and rationality, we therefore see the present account as an argument for the irreducibility of the meaning of vague expressions to those of precise expressions.\footnote{\cite{sutton2018probabilistic} recently argued that an adequate metasemantics for probabilistic treatment of vagueness is one in which vague expressions do not have truth-conditions proper, but default rules of use. Our account of the meaning of ``around'' maintains truth-conditions for ``around'', but {as just discussed they do not play any direct role, {and our model could even be directly formulated in terms of a fuzzy semantics}. In our model, the listener, when interpreting an ``around''-statement, updates her probability distribution over worlds, but does not exclude any world from the common ground}. We leave a more detailed discussion of this aspect, as well as of the rejoinders that could be made on behalf of epistemicism, for another occasion.} On Williamson's epistemicist perspective, vague expressions must be used with a margin of error to make sure that they are used truthfully. Here, a vague modifier like ``around'' is seen instead as a linguistic means to convey accurate information about one's state of uncertainty, and thereby as a resource allowing one to be maximally informative while still securing truthfulness.}

\section{Conclusion}\label{sec:conclusion}

In this paper we have pursued two main goals, one broad and one more specific. Our broad goal has been to flesh out the general idea that vague language can be more optimal than precise language in some contexts. One side to that idea is already epitomized in \cite{frazee2010vagueness}'s dictum that ``vagueness is rational under uncertainty", and in their proposal to substantiate this view in probabilistic and information-theoretical terms. However, another side to it is novel, namely the idea that vagueness can allow a cooperative speaker to achieve an optimal tradeoff between Gricean Quality and Gricean Quantity.

\bs{To establish this, we have shown that when a speaker is uncertain about the world, the use of a vague preposition like ``around'' offers {in some cases} an optimal way to secure Quality (truthfulness) and Quantity (informativeness). That is, we have shown that the use of ``around'' can be informationally optimal compared to any more precise way for the speaker to convey the information at her disposal (whether by means of exact numerical values or of precise intervals). 

Of course, we do not claim that our approach can be used to explain all the factors that can rationalize vagueness. We share the view that vagueness is a multi-source phenomenon, and that further constraints on learning and concept formation need to be taken into consideration.\footnote{See in particular \cite{franke2018vagueness} and \cite{douven2019rationality}.} 
What we hope to have shown is that an adequate account of linguistic vagueness is grounded in part in general pragmatic principles of successful communication, and, more specifically, that vague expressions make it possible for speakers to convey probabilistic information, in a way that precise expressions cannot (outside of explicit probability talk), and with no need to assume that their lexical entry directly refers to probabilities.

}

\newpage

\begin{appendices}

\let\oldarabic\arabic

\renewcommand{\arabic}[1]{A-\oldarabic{#1}}

\setcounter{ExNo}{0}


\section{A limitation result about the LU model}\label{app:LU}

{
\noindent In this appendix, we prove that in the lexical uncertainty model (LU model, \citealt{bergen2016pragmatic}), if two observations $o_1$ and $o_2$ are such that the support of the conditional distributions $P(w|o_1)$ and $P(w|o_2)$ are the same, then, at every level of the recursion, the speaker's probability of using a message $m$ if she observed $o_1$ is the same  as if she observed $o_2$. It follows that in the LU model, only the \emph{support} of the subjective probability distribution of the speaker, and not its \emph{shape}, plays a role in her choice of a message, in contrast with our model. \\
\\
The LU model is defined by the following equations, where $\sem{m}^i(w)$ is the truth-value of the literal meaning of $m$, relative to the interpretation function $i$, in world $w$, and $\sem{m}^i$ denotes the set of worlds where $m$ is true relative to interpretation $i$  (in our setting $i$ is what determines the interpretation of `around', i.e. a certain value for $y$). The parameter $\lambda$ is a non-null, positive real number. For any message $m$, $c(m)$ is the \emph{cost} of $m$, a null or positive real number. $P$ is the prior distribution about the possible values of $w$ (world state), $o$ (observation) and $i$, and is such that the value taken by $i$ is probabilistically independent from $w$ and $o$.

\begin{enumerate}[label=\color{blue}\theenumi.]

\item $L^0(w,o | m,i) = \dfrac{P(w,o) \times \sem{m}^i(w)}{P(\sem{m}^i)}$
\item $U^1(m|o,i) = (\sum\limits_{w}P(w | o) \times \log(L^0(w,o | m,i))) - c(m)$\\
 \label{u}
\item $S^1(m | o, i) = \dfrac{\exp(\lambda U^1(m,o,i))}{ \sum\limits_{m'}\exp(\lambda U^1(m',o,i))}$ \label{iii}
\item $L^1(w,o | m) = \dfrac{P(w,o)\times \sum\limits_{i} P(i)S^1(m | o, i))}{\alpha_1(m)}$, where $\alpha_1(m) = \sum\limits_{w',o'} (P(w',o')\sum\limits_{i} P(i)S^1(m | o', i))$ 
\item For $n \geq 1$, $U^{n+1}(m|o) = (\sum\limits_w P(w| o)\log(L^n(w,o | m)) - c(m)$
\item $S^{n+1}(m | o) = \dfrac{\exp(\lambda U^{n+1}(m,o))}{\sum\limits_{m'} \exp(\lambda U^{n+1}(m',o))}$ \label{vi}
\item For $n \geq 2$, $L^n(w,o | m) = \dfrac{P(w,o) \times S^n(m | o)}{\alpha_n(m)}$, where $\alpha_n(m) = \sum\limits_{w',o'}P(w',o')S^n(m | o')$


\end{enumerate}

\subsubsection*{Results to be proven}

If $o$ is an observation, $P_o$ is the probability distribution over worlds defined by:\\ $P_o(w) = P(w,o|o) = P(w|o)$

\begin{enumerate}
\item If two observations $o_1$ and $o_2$ are such that $P_{o_1}$ and $P_{o_2}$ have the same support (i.e. for every $w$, $P(w|o_1)>0$ iff $P(w|o_2) > 0$), then for every message $m$ and every interpretation function $i$, $S^1(m|o_1, i) = S^1(m|o_2, i)$
\item If two observations $o_1$ and $o_2$ are such that $P_{o_1}$ and $P_{o_2}$ have the same support, then for every $n \geq 2$, $S^n(m|o_1) = S^n(m|o_2)$
\end{enumerate}

\subsection{Proof strategy} \label{strat}

The key ingredient of the proof is the following. Consider two observations $o_1$ and $o_2$ such that $P_{o_1}$ and $P_{o_2}$ have the same support (i.e. they assign a non-null probability to the same worlds). Consider the set of messages $\mathcal{M}$ that express a proposition that is entailed by this support under at least one interpretation $i$ (a condition we call \emph{Weak Quality} -- as we shall see, other messages are not usable at all by a speaker who observed $o_1$ or $o_2$, as they violate Grice's maxim of Quality, cf. \ref{fact} below). We prove that, at every level of recursion, the utility achieved by each message in $\mathcal{M}$ for a speaker who observed $o_2$ is equal to the one achieved for a speaker who observed $o_1$ \textbf{plus a constant term which does not depend on the message}. In other words, there is a quantity $K$ which depends on the various parameters of the models and on $o_1$  and $o_2$ \emph{but crucially not on the message $m$},  such that for every message $m$, $U^n(m|o_2) = U^n(m|o_1) + K$. 

Now, the speaker strategy in the RSA model as defined in Equations \ref{iii}~and \ref{vi}\ is based on the so-called \emph{SoftMax} function.\footnote{In Section \ref{sm} we provide the definition of the SoftMax function and restate Equations  \ref{iii}~and \ref{vi} in terms of it.} The SoftMax function turns a sequence of numerical values (in our case the utilities of each message relative to a certain observation) into a probability distribution over members of this sequence. In the RSA model, the speaker's strategy relative to a given observation is obtained by applying the SoftMax function to the utilities of each message relative to this observation. 
The SoftMax function enjoys a property known as \emph{translation invariance}, as we prove below.\footnote{See also \url{https://en.wikipedia.org/wiki/Softmax_function\#Properties}} 
That is, if, starting from a sequence of numerical values, we consider the sequence obtained by adding a constant term to each value, and then apply the SoftMax function to these shifted values, the resulting probability distribution over the members of the new sequence is exactly the same as the one obtained when applying the SoftMax function to the initial sequence.
Since the utilities achieved by different messages relative to $o_2$ can be obtained by adding a constant term to those achieved by the very same messages relative to $o_1$ (as explained in the previous paragraph), this means that the resulting probability distribution over messages is the same for $o_1$ and $o_2$.

We will first prove that for $S^1$, relative to a fixed interpretation $i$, the difference between the utility achieved by a message $m$ in $\mathcal{M}$ relative to $o_1$ and the one achieved by the same message relative to $o_2$ does not depend on $m$, and is therefore the same across messages in $\mathcal{M}$. Thanks to translation invariance, this will ensure that for every $m$ in $\mathcal{M}$ and every $i$, $S^1(m|o_2, i) = S^1(m|o_1, i)$. Then we will do the same with $S^2$ -- there has to be a separate step for $S^2$ because the variable $i$ that appears in the definition of $S^1$ no longer appears when $n \geq 2$, so we start the inductive proof at level 2.  Then we complete the proof by proving the inductive step: assuming that for every $n \geq 2$, $S^n(m|o_2) = S^n(m|o_1)$, we prove that at the $n+1$-level, the difference in the utility achieved by a message $m$ in ${\mathcal M}$ relative to $o_2$ and relative to $o_1$ does not depend on $m$, which, thanks to translation invariance, ensures that for every $m$ in ${\mathcal M}$, $S^{n+1}(m|o_2) = S^{n+1}(m|o_1)$. At each step, the reason why the key result is observed is that the log-function which appears in the definition of the utility function turns products into sums, allowing us, together with the fact that probabilities sum up to 1,  to separate terms which depend on $m$ from those that depend on $o$, and all terms in which $m$ appears cancel out when we consider the difference $U(m|o_2)-U(m|o_1)$.\\
\\
We now give the detailed proof.

\subsection{Quality and Weak Quality}

\paragraph{Definitions}
\begin{enumerate}
\item We say that a message $m$ respects Quality with respect to an observation $o$ and an interpretation $i$ if, for every $w$, if  $P(w | o) > 0$, then $\sem{m}^i(w)=1$.
\item We say that a message $m$ respects Weak Quality with respect to an observation $o$ if there exists an interpretation $i$ such that $P(i)>0$ and $m$ respects Quality with respect to $o$ and $i$.
\end{enumerate}

\noindent Let $o_1$ and $o_2$ be two observations such that $P_{o_1}$ and $P_{o_2}$ have the same support, i.e. for every $w$, $P(w|o_1)>0$ iff $P(w|o_2) > 0$. From now on, $o_1$ and $o_2$ denote two such observations.

\ex. \textbf{Lemma} \label{el}
\a. A message $m$ respects Quality with respect to $o_1$ and some interpretation $i$ if and only if $m$ respects Quality with respect to $o_2$ and the same interpretation $i$.
\b. A message $m$ respects Weak Quality with respect to $o_1$ if and only if it respects Weak Quality with respect to $o_2$.

\subsubsection*{Proof of Lemma \Last}

Obvious from the definitions of Quality and Weak Quality, and the fact that $P_{o_1}$ and $P_{o_2}$ have the same support.

\ex. \textbf{Facts}.\label{fact}
\a. If a message $m$ does not respect Quality with respect to an observation $o$ and an interpretation $i$, then $U^1(m|o,i) = -\infty$ and $S^1(m|o,i)=0$; if $m$ does respect Quality with respect to $o$ and $i$, then $U^1(m|o,i) \neq -\infty$ and  $S^1(m|o,i)>0$ \label{fa}
\b.If a message $m$ does not respect Weak Quality with respect to an observation $o$, then for every $n \geq 2$, $U^n(m|o) = -\infty$ and  $S^n(m|o)=0$. If $m$ does respect Weak Quality with respect to $o$, then $U^n(m|o) \neq -\infty$ and  $S^n(m|o)>0$. \label{fb}

\subsubsection*{Proof of the facts in \Last}
First we prove the result for $S^1$, then for $S^2$ and then by induction for higher values of $n$.\\
\\
\textbf{Proof of \ref{fa}} --  If $m$ does not respect Quality with respect to $o$ and $i$, then for some $w$ such that $P(w|o) > 0$, $\sem{m}^i(w)=0$. For such a $w$, then, $L^0(w,o|m,i)=0$, hence $\log(L^0(w,o|m,i)) = - \infty$. So at least one term in the sum which defines $U^1(m|o,i)$ evaluates to $-\infty$, and so the sum itself does, hence $U^1(m|o,i) = - \infty$.\footnote{{Strictly speaking, of course, $U^1(m|o,i)$ is simply not defined, since $\log(0)$ is not defined. The point is simply that the limit of $\exp(f(x))$ in 0 is 0 when $f$ diverges to  $-\infty$ in 0. Likewise, we also treat the function $x\times \log(x)$ as evaluating to $0$ in $0$, because even though this function is not defined in $0$, its limit in $0$ is $0$. Here and elsewhere we choose not to introduce explicit reasoning about limits in order to simplify the exposition, with no harmful effects.}}
  Since $S^1(m|o,i)$  involves exponentiating a quantity that is infinitely negative, we have $S^1(m|o,i) = 0$.~Reciprocally, if $m$ does respect Quality with respect to $o$ and $i$, then no term in the sum is infinitely negative, and $U^1(m|o,i)$ is not infinitely negative either, and so $S^1(m|o,i)>0$\\
\\
\textbf{Proof of \ref{fb}} --  The proof is by induction.\\
Base-case ($n=2$): Suppose now that $m$ does not respect Weak Quality with respect to $o$. Then for every $i$ such that $P(i)>0$, $m$ does not respect Quality with respect to $o$, $i$; and so by the result just proven, every term in the sum $\sum\limits_{i} P(i)S^1(m | o, i)$ is equal to 0, and so is the sum as a whole. As a result, $L^1(w,o|m)=0$, for every $w$. From this it follows that the sum $\sum\limits_w P(w| o)\log(L^1(w,o | m))$ evaluates to $-\infty$, and therefore so does $U^2(m|o)$. $S^2(m|o)$ involves again exponentiating an infinitely negative value, so is equal to 0. Reciprocally, if $m$ respects Weak Quality with respect to $o$, there is at least one $i$ relative to which $P(i)\times S^1(m | o, i)>0$, and therefore $\sum\limits_{i} P(i)S^1(m | o, i)$ is not equal to $0$. For every $w$ such that $P(w,o)>0$, then $L^1(w,o|m)>0$, and therefore $U^2(m|o)$ is not infinitely negative, and so $S^2(m|o)>0$.\\
\\
Inductive step: Finally, let assume that the result holds for $S^n$ (Induction Hypothesis). Suppose again that $m$ does not respect Weak Quality with respect to $o$. By the Induction Hypothesis, $S^n(m|o)=0$, and therefore for every $w$, $L^n(w,o|m)=0$ (given the definition of $L^n$). Then $U^{n+1}(m|o) = - \infty$, as in each term of the sum that defines $U^{n+1}(m|o)$, the $\log$-function takes 0 as its argument. It follows that $S^{n+1}(m|o)=0$. Reciprocally, if $m$ does respect Weak Quality with respect to $o$, then $S^n(m|o)>0$, and therefore for some $w$ ($w$ must be such that $P(w,o) > 0$), $L^n(w,o|m)>0$, from which it follows that $U^{n+1}(m|o)$ is not infinitely negative and therefore that $S^{n+1}(m|o) >0$.

\subsection{Reformulating Utility Functions in terms of SoftMax, SoftMax invariance}

\subsubsection*{The SoftMax Function} \label{sm}

The \textbf{SoftMax} function takes three arguments: a sequence of real numbers $\vec{x} = (x_1, \ldots, x_i)$, a member of this sequence, and the paratemer $\lambda$. It is defined as follows:
$$\textrm{SoftMax}(x_k, \vec{x}, \lambda) = \dfrac{\exp(\lambda x_k)}{\sum\limits_{x_i \in \vec{x}} \exp(\lambda x_i)}$$

\subsubsection*{Reformulating the Utility function in terms of SoftMax}

Equation \ref{iii} is repeated here:
$$S^1(m | o, i) = \dfrac{\exp(\lambda U^1(m,o,i))}{ \sum\limits_{m'}\exp(\lambda U^1(m',o,i))}$$
\noindent Note that in the denominator, every term of the sum corresponding to a message which does not respect Weak Quality with respect to $o$ is equal to $0$ (because its utility is infinitely negative, and so after exponentiation we get $0$, cf. \ref{fa}). This means that we can restrict the denominator to the messages that respect Weak Quality with respect to $o$. \\Let $\mathcal{M}_o = <m_1, \ldots, m_j, \ldots>$ be an ordered sequence of messages which contains all and only the messages that respect Quality with respect to $o$.\footnote{To properly define this sequence, we choose once and for all an enumeration of all messages, and we order the sequence in accordance with this enumeration.} The above Equation can then be rewritten as:
$$S^1(m | o, i) = \dfrac{\exp(\lambda U^1(m,o,i))}{ \sum\limits_{m_j \in \mathcal{M}_o}\exp(\lambda U^1(m_j,o,i))}$$
\noindent Let $\overrightarrow{U^1_{o,i}}$ be be the sequence of numbers that one gets by applying the function $U^1(\ldots|o,i)$ pointwise to the sequence $\mathcal{M}_o$ (i.e. $\overrightarrow{U^1_{o,i}}=  <U^1(m_1|o, i), \ldots, U^1(m_j|o,i), \ldots>$)
\\
\noindent Then Equation \ref{iii} can be rewritten as follows:

\ex.$S^1(m | o, i) = \textrm{SoftMax}(U^1(m|o,i), \overrightarrow{U^1_{o,i}}, \lambda)$ \label{sm1}

Consider now Equation \ref{vi}, which we repeat here: 
 $$S^{n+1}(m | o) = \dfrac{\exp(\lambda U^{n+1}(m|o))}{\sum\limits_{m'} \exp(\lambda U^{n+1}(m'|o))}$$
\noindent For any $n$, $U^{n+1}(m'|o) = -\infty$ if $m'$ does not respect Weak Quality with respect to $o$ (cf. \ref{fb}), and so every term in the sum that constitutes the denominator that corresponds to such a message $m'$ is equal to $0$ (exponentiation of $-\infty$). Hence we can again restrict the sum to members of $\mathcal{M}_o$. With $\overrightarrow{U^{n+1}_o} = <U^{n+1}(m_1|o), \ldots, U^{n+1}(m_j|o), \ldots>$,  where $< \ldots, m_j, \ldots>$ is a sequence in which all and only members of $\mathcal{M}_o$ appear, we get:

\ex. For any $n \geq 1$, $S^{n+1}(m | o) = \textrm{SoftMax}(U^{n+1}(m), \overrightarrow{U^{n+1}_o},\lambda)$ \label{sm2}

\subsubsection*{Translation Invariance of SoftMax}

\textbf{Notation:}~  If $\vec{x}$ is a sequence of real numbers  $<x_1, \ldots, x_i, \ldots>$ and $a$ is a real number, we notate $\vec{x} \oplus a$ the sequence $<x_1 +a,\ldots, x_i + a, \ldots>$.

\ex. \textbf{Lemma} : $\textrm{SoftMax}$ is translation invariant.\\
 If $\vec{x}$ is a sequence of real numbers, $x_k$ a member of of $\vec{x}$ and $a$ a  real number and $\lambda$ is a positive real number, $\textrm{SoftMax}(x_k + a, \vec{x} \oplus a, \lambda) = \textrm{SoftMax}(x_k, \vec{x}, \lambda)$ \label{trans}

\textbf{Proof}
\begin{eqnarray*}
\textrm{SoftMax}(x_k + a, \vec{x} \oplus a, \lambda) 
 &=& \dfrac{\exp(\lambda x_k + a)}{\sum\limits_{y_j \in \vec{x} \oplus a} \exp(\lambda y_j)}\\
 &=& \dfrac{\exp(\lambda x_k + a)}{\sum\limits_{x_j \in \vec{x}} \exp(\lambda (x_j + a))}\\
&=& \dfrac{\exp(\lambda x_k) \times \exp(\lambda a)}{\sum\limits_{{x_j} \in \vec{x}} \exp(\lambda x_j) \times \exp(\lambda a)}\\
&=& \text{\footnotesize{($\exp(\lambda a)$ simplifies)}}\\
& & \dfrac{\exp(\lambda x_k)}{\sum\limits_{{x_j} \in \vec{x}} \exp(\lambda x_j)}\\
&=& \textrm{SoftMax}(x_k, \vec{x}, \lambda) 
\end{eqnarray*}

\subsection{Proving the result for $S^1$}

Recall that $o_1$ and $o_2$ are two observations such that $P_{o_1}$ and $P_{o_2}$ have the same support.

\ex. \textbf{Core Lemma} \label{Lemma1}\\
If $m$ respects Quality, relative to $i$, with respect to $o_1$ and $o_2$, then the difference $U^1(m | o_2, i) - U^1(m | o_1, i)$ does not depend on $m$ or $i$, but only on $o_1$ and $o_2$ (i.e. it is the same for any $m$ that respects Quality with respect to $o_1$ and $o_2$, and $i$).  More specifically, with $S$ being the support of $P_{o_1}$ and $P_{o_2}$: $$U^1(m | o_2, i) - U^1(m | o_1, i) =  \sum\limits_{w \in S}P(w | o_2)\log(P(w,o_2)) - \sum\limits_{w \in S}P(w | o_1)\log(P(w,o_1))$$ ($m$ and $i$ do not appear on the right-hand side).
\subsubsection*{Proof of the Lemma in \ref{Lemma1}}

Let us assume that $m$, $i$ and $o_1$ and $o_2$ meet the condition stated in \ref{Lemma1}, i.e. that $m$ respects Quality with respect to $i$ and $o_1$, and with respect to $i$ and $o_2$.
\begin{eqnarray*}
U^1(m | o_1, i) & =& \sum\limits_{w}P(w | o_1)\log(L^0(w,o_1 | m,i)) - c(m)\\
&=& \sum\limits_{w}P(w | o_1)\log \left(\dfrac{P(w,o_1)\times\sem{m}^i(w)}{P(\sem{m}^i)}\right) - c(m)\\
\end{eqnarray*}
Let us notate $S$ the support of $P_{o_1}$ and $P_{o_2}$. Note that $P(w|o_1) = 0$ if $w \notin S$. Furthemore, since $m$ respects Quality with respect to $o_1$, $o_2$, and $i$, then if $w \in S$, $\sem{m}^i(w) = 1$. It follows that we can restrict the sum to the worlds in $S$ (because all the terms in the sum are equal to $0$ when $w$ is not in $S$), and that, having done this, we can remove $\sem{m}^i(w)$ from the equation (since $\sem{m}^i(w)$  is always equal to $1$ when $w \in S$). We can therefore continue as follows:
\begin{eqnarray*}
U^1(m | o_1, i) &= &\sum\limits_{w \in S}P(w | o_1)\log \left(\dfrac{P(w,o_1)}{P(\sem{m}^i)}\right) - c(m) \\
&=&  \sum\limits_{w \in S}P(w | o_1)[\log(P(w,o_1) - \log(P(\sem{m}^i))] - c(m)\\
&=& \sum\limits_{w \in S}P(w | o_1)\log(P(w,o_1)) - \sum\limits_{w \in S}P(w | o_1)\log(P(\sem{m}^i)) - c(m)\\
&=&\sum\limits_{w \in S}P(w | o_1)\log(P(w,o_1)) - \log(P(\sem{m}^i))\times \underbrace{\sum\limits_{w \in S}P(w | o_1)}_{=1} - c(m)\\
&=& \sum\limits_{w \in S}P(w | o_1)\log(P(w,o_1)) - \log(P(\sem{m}^i)) - c(m).
\end{eqnarray*}
\noindent The same formula of course holds for $o_2$, replacing every occurrence of $o_1$ with $o_2$. Given this, when we substract $U^1(m | o_1, i)$ from $U^1(m | o_2, i)$, the terms that depend on $m$ ($- \log(P(\sem{m}^i)) - c(m)$) cancel out, and we get: 
\begin{eqnarray*}
 U^1(m | o_2, i) - U^1(m | o_1, i) =  \sum\limits_{w \in S}P(w | o_2)\log(P(w,o_2)) - \sum\limits_{w \in S}P(w | o_1)\log(P(w,o_1))
\end{eqnarray*}
As promised, then, this difference does not depend on $m$ or $i$.
\\
\ex. \textbf{Theorem}\\
If two observations $o_1$ and $o_2$ are such that the distributions (over $w$) $P_{o_1}$ and $P_{o_2}$ have the same support, then, for every interpretation $i$ and every message $m$, $S^1(m | o_1,i) = S^1(m | o_2,i)$.  \label{2}

\subsubsection*{Proof of \ref{2}}

First consider the case where $m$ does not respect Quality with respect to $o_1$, $o_2$, $i$ (again, relative to a fixed $i$, either it respects Quality for both $o_1$ and $o_2$, or for neither, cf. \ref{el}). In the case, given the first fact in \ref{fact}, $S^1(m | o_1, i) = S^1(m | o_2, i) = 0$.\\
\\
Consider now the case were $m$ respects Quality with respect to $o_1$, $o_2$, $i$. Let us define $K(o_1, o_2) =  \sum\limits_{w \in S}P(w | o_2)\log(P(w,o_2)) - \sum\limits_{w \in S}P(w | o_1)\log(P(w,o_1))$. From the lemma in \ref{Lemma1}, we have: for every $m'$ which respects Quality with respect to $o_1$, $o_2$, $i$, $$U^1(m' | o_2, i) =U^1(m' | o_1, i) + K(o_1, o_2).$$

\noindent Given that this result holds for all the messages that respect Quality with respect to $o_1$, $o_2$ and $i$ (recall that the messages that respect Quality w.r.t. $o_1$, $i$ are the same as those that respect it w.r.t. $o_2$, $i$),  it can be restated as follows, using the notations introduced in Section \ref{sm}:\footnote{Recall that $\overrightarrow{U^1_{o,i}}$ is a sequence that contains all the utilities achieved by messages that respect Quality with respect to $o$ and $i$, ordered according to a fixed enumeration of all messages.}:
$$\overrightarrow{U^1_{o_2,i}} = \overrightarrow{U^1_{o_1,i}} \oplus K(o_1,o_2)$$
\noindent Given \ref{sm1} and Translation Invariance (\ref{trans}), we have:
\begin{eqnarray*}
S^1(m | o_2, i)& = & \textrm{SoftMax}(U^1(m|o_2, i), \overrightarrow{U^1_{o_2,i}},\lambda)\\
&=& \textrm{SoftMax}(U^1(m|o_1, i) + K(o_1, o_2),  \overrightarrow{U^1_{o_1,i}} \oplus K(o_1,o_2),\lambda)\\
&=& (\textrm{{\small Translation Invariance}})~   \textrm{SoftMax}(U^1(m|o_1, i),  \overrightarrow{U^1_{o_1,i}},\lambda)\\
&=& S^1(m|o_1, i)
\end{eqnarray*}

\subsection{Proving the result for $n \geq 2$}

\ex. \label{th} \textbf{Theorem}\\
Let $o_1$ and $o_2$ be such that $P_{o_1}$ and $P_{o_2}$ have the same support. Then, for any $n \geq 2$, and any message $m$, $S^n(m | o_2)~=~S^n(m | o_1)$

This will be a proof by induction.\\
\\
Recall again that $o_1$ and $o_2$ are two observations such that $P_{o_1}$ and $P_{o_2}$ have the same support.\\
\\
First, note that if a certain message $m$ does not satisfy Weak Quality with respect to $o_1$, $o_2$, then given the second fact in \ref{fact}, for any $n \geq 2$, $S^n(m|o_1)= S^n(m|o_2)=0$, so we can now ignore this case and assume for the rest of the proof that $m$ does satisfy Weak Quality with respect to $o_1$ and $o_2$ (recall that it either respects it for both or neither, cf. Lemma \ref{el}).

\subsubsection{Base-Case: n =2}

We start with a counterpart to the Lemma in \ref{Lemma1}.

\ex.  \textbf{~~Lemma}\\
If $m$ respects Weak Quality relative to both $o_1$ and $o_2$, then the difference $U^2(m | o_2)~-~U^2(m | o_1)$ does not depend on $m$, but only on $o_1$ and $o_2$ (i.e. it is the same for any $m$ that respects Weak Quality with respect to $o_1$ and $o_2$).\\More specifically, with $S$ defined as the support of $P_{o_1}$ and $P_{o_2}$ : $$U^2(m | o_2)~-~U^2(m | o_1)~= ~\sum\limits_{w \in S}P(w | o_2)\log(P(w,o_2)) - \sum\limits_{w \in S}P(w | o_1)\log(P(w,o_1))$$ ($m$ does not appear on the right-hand side). \label{lemma X}

\subsubsection*{Proof of \ref{lemma X}}
Assume that $m$, $o_1$ and $o_2$ meet the condition of the above Lemma. Note that, since $P_{o_1}$ and $P_{o_2}$ have the same support, the interpretations $i$ such that  $m$ respects Weak Quality with respect to $o_1$ and $i$ are the same as the interpretations $i$ such that $m$ respects Weak Quality with respect to $o_2$  and $i$.(cf. Lemma \ref{el}). As before, we call $S$ the support of $P_{o_1}$ and $P_{o_2}$.
\\
We have, given the definitions:\footnote{{We can restrict the sum that defines $U^2(m | o_1)$  and $U^2(m | o_2)$ to the worlds of $S$, i.e. the support of $P_{o_1}$ and $P_{o_2}$, because for worlds $w$ outside of $S$, $P(w|o_1) = P(w|o_2) = 0$, $P(w,o_1) = P(w,o_2) = 0$, and so the corresponding terms in the sum are of the forms $0 \times log(0)$, i.e. $0$ -- since the limit of $x \rightarrow x\times log(x)$ in $0$ is $0$}} 
\begin{eqnarray*}
U^2(m | o_1) &=& \sum\limits_{w \in S} P(w | o_1) \times \log \left(\frac{P(w,o_1)\sum\limits_{i}P(i)S^1(m | o_1, i)} {\alpha_1(m)} \right ) \\
&=& \sum\limits_{w \in S} P(w | o_1) \times [\log(P(w,o_1)) + \log \left(\sum\limits_{i}P(i)S^1(m | o_1, i)\right) - \log(\alpha_1(m))]
\end{eqnarray*}

Likewise, we have:
\begin{eqnarray*}
U^2(m | o_2) = \sum\limits_{w \in S} P(w | o_2) \times [\log(P(w,o_2)) + \log \left(\sum\limits_{i}P(i)S^1(m | o_2, i)\right) - \log(\alpha_1(m))]
\end{eqnarray*}
Recall that for every $i$, $S^1(m | o_1,i) = S^1(m | o_2, i)$ (Theorem in \ref{2}). It follows that the quantities  $\log(\sum\limits_{i}P(i)S^1(m | o_1, i))$ and  $\log(\sum\limits_{i}P(i)S^1(m | o_2, i))$ are equal. Let us call this quantity $X$. We can rewrite the above formulae as:
\begin{eqnarray*}
U^2(m | o_1) = \sum\limits_{w \in S} P(w | o_1) \times [\log(P(w,o_1)) + X - \log(\alpha_1(m))]\\
U^2(m | o_2) = \sum\limits_{w \in S} P(w | o_2) \times [\log(P(w,o_2)) + X - \log(\alpha_1(m))]
\end{eqnarray*}
A few lines of computation yields the lemma stated in \ref{lemma X}:
\begin{eqnarray*}
U^2(m | o_2) -  U^2(m | o_1) &=&
(X - \log(\alpha_1(m))\times \underbrace{\left(\underbrace{\sum\limits_{w \in S} P(w | o_2)}_{=1} - \underbrace{\sum\limits_{w \in S}P(w | o_1)}_{=1}\right)}_{=0}\\
&+& \sum\limits_{w \in S} P(w | o_2)\log(P(w,o_2)) - \sum\limits_{w \in S} P(w | o_1)\log(P(w,o_1))\\
&=& \sum\limits_{w \in S} P(w | o_2)\log(P(w,o_2)) - \sum\limits_{w \in S} P(w | o_1)\log(P(w,o_1))
\end{eqnarray*}
%
\noindent As promised, this difference does not depend on $m$.

\subsubsection*{Using translation invariance to prove the base-case (n=2)}

To complete the proof for the base case ($S^2(m|o_1) = S^2(m|o_2)$), we just need to exploit again the translation-invariance property of SoftMax. The computation proceeds in exactly the same way as in the proof of the Theorem in \ref{2}: 
%
%
%
%

\noindent With $K(o_1, o_2) = \sum\limits_{w \in S} P(w | o_2)\log(P(w,o_2)) - \sum\limits_{w \in S} P(w | o_1)\log(P(w,o_1))$, we have, for any message $m$ that respects Weak Quality with respect to $o_1$, $o_2$, $U^2(m|o_2) = U^2(m|o_1) + K(o_1, o_2)$.\\
\\
Since the messages that respect Weak Quality with respect to $o_1$ and with respect to $o_2$ are the same, we have:
$$\overrightarrow{U^2_{o_2}} = \overrightarrow{U^2_{o_1}} + K(o_1, o_2)$$\\
We therefore have, thanks to Translation Invariance and \ref{sm2}:
%
\begin{eqnarray*}
S^2(m |o_2)& = & \textrm{SoftMax}(U^2(m|o_2), \overrightarrow{U^2_{o_2}},\lambda)\\
&=& \textrm{SoftMax}(U^2(m|o_1) + K(o_1, o_2),  \overrightarrow{U^2_{o_1}} \oplus K(o_1,o_2),\lambda)\\
&=& (\textrm{{\small Translation Invariance}})~   \textrm{SoftMax}(U^2(m|o_1),  \overrightarrow{U^2_{o_1}},\lambda)\\
&=& S^2(m|o_1)
\end{eqnarray*}

%
%
%

\subsubsection{Inductive step for the Theorem in \ref{th}}

Recall again that $P_{o_1}$ and $P_{o_2}$ have the same support.

\paragraph{Induction Hypothesis:} We assume that $S^n(m | o_2) = S^n(m | o_1)$.\\
\\
We want to prove that $S^{n+1}(m| o_2) = S^{n+1}(m | o_1)$.\\
\\
As before, the key intermediate result is the following:

\ex. \label{intfinal} \textbf{~~Intermediate result}\\
If $m$ respects Weak Quality relative to both $o_1$ and $o_2$ (and given the induction hypothesis), then the difference $U^{n+1}(m | o_2) - U^{n+1}(m | o_1)$ does not depend on $m$, but only on $o_1$ and $o_2$ (i.e. it is the same for any $m$ that respects Weak Quality with respect to $o_1$ and $o_2$). Again, with $S$ being the support of $P_{o_1}$ and $P_{o_2}$, we have:$$U^{n+1}(m | o_2) - U^{n+1}(m | o_1) =  \sum\limits_{w \in S}P(w | o_2)\log(P(w,o_2)) - \sum\limits_{w \in S}P(w | o_1)\log(P(w,o_1))$$ ($m$ does not appear on the right-hand side).

\subsubsection*{Proof of \ref{intfinal}}

We have:
\begin{eqnarray*}
U^{n+1}(m | o_1) &=& \sum\limits_{w \in S} P(w | o_1) \times \log \left(\dfrac{P(w,o_1)S^n(m | o_1)}{\alpha_n(m)}\right)\\
&=& \sum\limits_{w \in S} P(w | o_1) \times [\log(P(w,o_1)) + \log(S^n(m | o_1)) - \log(\alpha_n(m))]
\end{eqnarray*}
\noindent Likewise, we have:
\begin{eqnarray*}
U^{n+1}(m | o_2) &=&\sum\limits_{w \in S} P(w | o_2) \times [\log(P(w,o_2) + \log(S^n(m | o_2)) - \log(\alpha_n(m))]
\end{eqnarray*}
By the induction hypothesis, $\log(S^n(m | o_2)) = \log(S^n(m | o_1))$. Calling this quantity $X$, we then have:
\begin{eqnarray*}
U^{n+1}(m | o_1) &=& \sum\limits_{w \in S} P(w | o_1) \times [\log(P(w,o_1)) + X - \log(\alpha_n(m))]\\
U^{n+1}(m | o_2) &=& \sum\limits_{w \in S} P(w | o_2) \times [\log(P(w,o_2)) + X - \log(\alpha_n(m))]\text{,}
\end{eqnarray*}

\noindent 
The computation then proceeds exactly as in the proof for the Lemma in \ref{lemma X} -- terms that depend on $m$ cancel out when we take the difference between the two lines, and we so we can conclude the proof of \ref{intfinal}: $$U^{n+1}(m | o_2) - U^{n+1}(m | o_1) =  \sum\limits_{w \in S}P(w | o_2)\log(P(w,o_2)) - \sum\limits_{w \in S}P(w | o_1)\log(P(w,o_1))$$

\subsubsection*{Completing the proof using Translation Invariance}

On the basis of \ref{intfinal}, the proof that $S^{n+1}(m|o_2) = S^{n+1}(m|o_1)$ proceeds in exactly the same way as in the case of $n=2$, and boils down to translation invariance again.\\
\\
With $K(o_1, o_2) = \sum\limits_{w \in S} P(w | o_2)\log(P(w,o_2)) - \sum\limits_{w \in S} P(w | o_1)\log(P(w,o_1))$, we have, for any message $m$ that respects Weak Quality with respect to $o_1$, $o_2$, $U^{n+1}(m|o_2) = U^{n+1}(m|o_1) + K(o_1, o_2)$.\\
\\
Since the messages that respect Weak Quality with respect to $o_1$ and with respect to $o_2$ are the same, we have:

$$\overrightarrow{U^{n+1}_{o_2}} =  \overrightarrow{U^{n+1}_{o_1}} \oplus K(o_1, o_2)$$

\noindent Given Translation Invariance and \ref{sm2}, we have:
\begin{eqnarray*}
S^{n+1}(m |o_2)& = & \textrm{SoftMax}(U^{n+1}(m|o_2), \overrightarrow{U^{n+1}_{o_2}},\lambda)\\
&=& \textrm{SoftMax}(U^{n+1}(m|o_1) + K(o_1, o_2),  \overrightarrow{U^{n+1}_{o_1}} \oplus K(o_1,o_2),\lambda)\\
&=& (\textrm{{\small Translation Invariance}})~   \textrm{SoftMax}(U^{n+1}(m|o_1),  \overrightarrow{U^{n+1}_{o_1}},\lambda)\\
&=& S^{n+1}(m|o_1)
\end{eqnarray*}
%
\noindent This completes the proof of \ref{th}.
}


\section{An alternative model for the literal listener}\label{app:original}

The model presented in section \ref{sec:semaround} was originally derived from a distinct model of the listener that we first came up with, which we present in this appendix for comparison. Although that model makes basically the same qualitative predictions,  it is not a Bayesian model, and it makes different quantitative predictions. 

Like the Bayesian model, this model assumes that the listener has a probability distribution $P$ over the intervals selected by ``around'', and over the possible values the variable $x$ of interest might have. However, the alternative model (\PE{written WIR, for Weighted Interpretation Rule}) says that the listener's posterior value of $x$ upon hearing ``$x$ is around $n$'', {notated $L(x = \ldots|\ x\ \textrm{is around}\ n)$}, is the sum of the conditional probabilities that $x$ takes that value when $x$ belongs to a given interval, weighted by the probability of that interval:\footnote{{This characterization of the listener happens to be identical (modulo differences in notations) to one that is discussed in Appendix B of \citet{bergen2016pragmatic}, statement (41).}}

\begin{equation}
{L}(x=k\ |\ x\ \textrm{is around}\ n)=_{df}\sum_{i} P(x=k\ |\ x\in [n-i, n+i])P(y=i)\tag{\textcolor{blue}{WIR}}\label{eq:orig_around}
\end{equation}

To see the difference with the Bayesian model, recall {Equation \ref{eq:around2}, which is reproduced here (with an explicit formula for the proportionality factor):

\begin{equation*}
P(x=k\mid \text{$x$ is around $n$}) = \dfrac{P(x=k) \times \sum\limits_{i \geq |n-k|} P(y=i)}{\sum\limits_k P(x=k)\times \sum\limits_{i \geq |n-k|} P(y=i)}\end{equation*}

}

The two models are distinct. For instance, when $P$ is uniform over values of $x$ as well as over candidate meanings for ``around'', the distribution {$L(x = \ldots|\ x\ \textrm{is around}\ n)$} is distinct from the one depicted in Figure \ref{fig:around20}, and it is no longer linear, as represented in Figure \ref{fig:orig_20}. 
\begin{figure}[t]
\[
\includegraphics[scale=0.4, trim= 0 1cm 1cm 1cm, angle = -90]{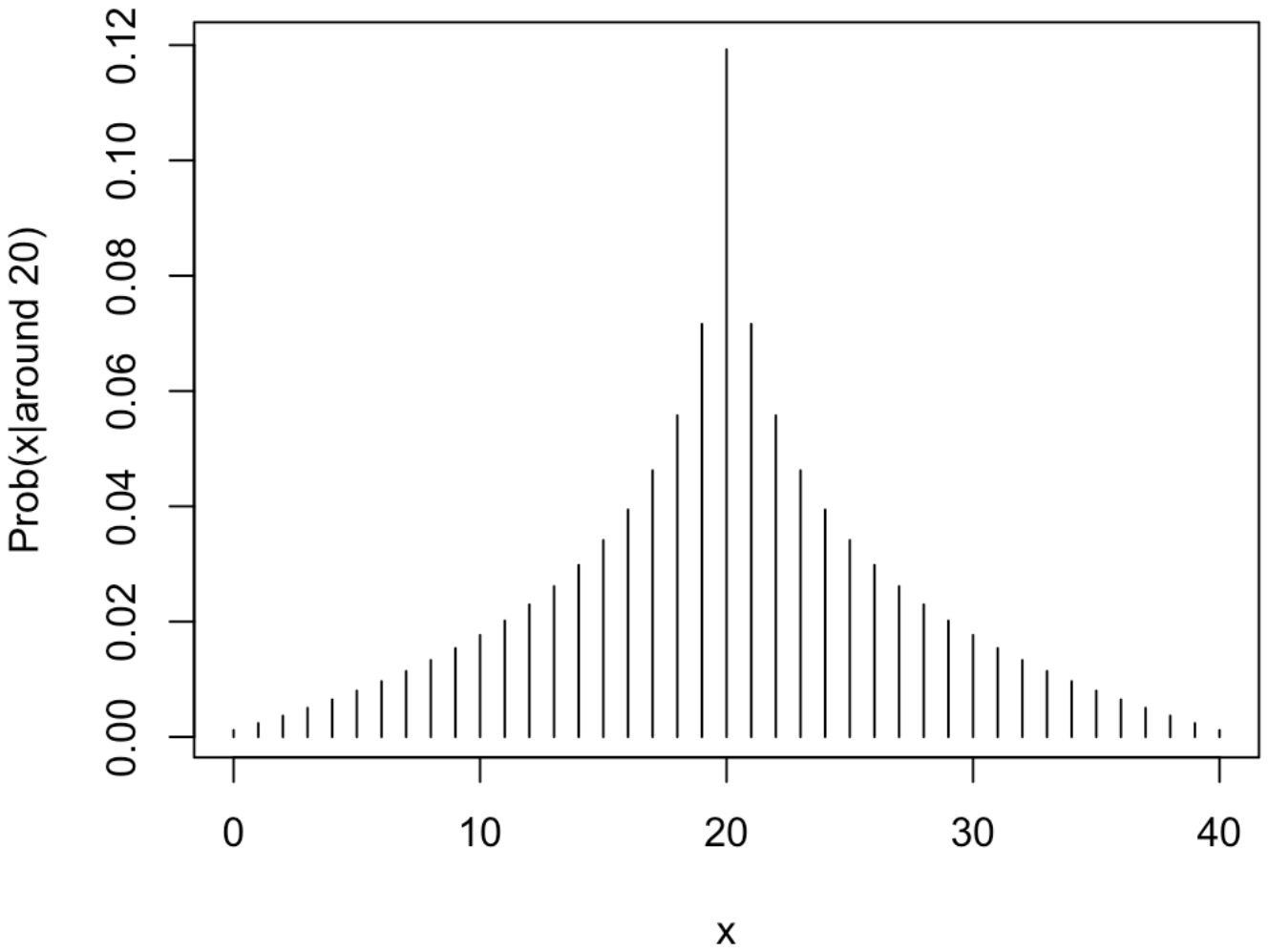}
\]
\caption{{$L(x=k|\ x\ \textrm{is around}\ n)$}, with maximum interval $[0,40]$}\label{fig:orig_20}
\end{figure}

{To see more precisely how the two models differ conceptually, the following observation will be useful.} Let $P_{post}$ be the posterior probability distribution resulting from updating $P$ with the ``around $n$'' message in the Bayesian model, i.e.: $$P_{post}(x=k, y=i) =_{df} P(x=k, y=i\ |\ x \in [n-y, n+y]).$$ It can be proved that:

\begin{equation}\label{eq:post} P_{post}(x=k) = \sum\limits_{i} P(x=k\ |\ x \in [n-i, n+i])P_{post}(y=i) \tag{\textcolor{blue}{BIR'}}
\end{equation}

Equation \ref{eq:orig_around} looks almost like \ref{eq:post}, except that the first term in \ref{eq:orig_around} is weighted by the \emph{prior} distribution on the values of $y$ (the candidate meanings for ``around'') instead of the posterior. This is the sense in which the model proposed in \ref{eq:orig_around} is not Bayesian. Instead of the listener updating also her probability of intervals after hearing ``$x$ is around $n$'', the listener does not make her interval probability depend on that information. The model is not illegitimate for that matter. Regarding our explananda, it makes the same central prediction: this model can be used to derive the Ratio Inequality. {If we used this model in order to characterize the literal Listener, we could still build an RSA model, in the same way as we did in section \ref{RSAmodel}, and we would derive qualitatively similar predictions.}

The proof of \ref{eq:post} goes as follows. $P_{post}$ being a probability distribution, it satisfies, for any $k$:

\begin{equation*}
P_{post}(x=k) = \sum\limits_{i}[P_{post}(x=k\ |\ y=i)\times P_{post}(y=i)]
\end{equation*}

Let us develop the first factor in the sum, $P_{post}(x=k\ |\ y=i)$.\\
Since, in general, $P_C(A|B) = P(A|B\wedge C)$ [where $P_C$ is $P$ conditionalized on event $C$] we have:
\begin{align*}
P_{post}(x=k\ |\ y=i) & = P(x=k\ |\ y=i \wedge x \in [n-y, n+y])\\
& = \dfrac{P(x=k \wedge y=i \wedge x \in [n-y, n+y])}{P(y=i \wedge x \in [n-y, n+y])}\\
& = \dfrac{P(x=k \wedge y=i \wedge x \in [n-i, n+i])}{P(y=i \wedge x \in [n-i, n+i])}
\end{align*}

Since the random variables $x$ and $y$ are independent, the events $\ulcorner y=i \urcorner$ and $\ulcorner x=k \wedge x \in [n-i, n+i] \urcorner$ are independent, thanks to which we can simplify the formula above:

\begin{align*}
P_{post}(x=k\ |\ y=i) & = \dfrac{P(y=i)\times P(x=k \wedge x \in [n-i, n+i])}{P(y=i) \times P(x \in [n-i, n+i])}\\
& = \dfrac{P(x=k \wedge x \in [n-i, n+i])}{P(x \in [n-i, n+i])}\\
& = P(x=k\ |\ x \in [n-i, n+i])
\end{align*}

Plugging this equality into the sum above, we get exactly \ref{eq:post}.


\section{Two alternative RSA models (discussed in section \ref{sec:comparison})} \label{app:variants}

\subsection{A variant of our model which uses the standard utility function} \label{sec:variant_our}

In our official model, the utility function for the speaker is defined by the following equations, where $P_o$ is understood to be the posterior distribution over the variable of interest (here notated $w$, for \textit{world}) induced by observation $o$, and $L^n_m$ is the posterior distribution over $w$ of the level-$n$ listener who has processed a message $m$. These distributions, importantly, are not joint distributions over $w$ and $o$.

$$U^{n+1}(m,o_j) =-D_{KL}(P_{o}||L^n_m)$$

Developing the formula for KL-divergence, this is more explicitly cashed out as:

\begin{align*}
U^n(m,o) =   \sum\limits_w P(w|o) \times [\log(L^n(w|m)) - \log(P(w|o))]\\
= \sum\limits_w P(w|o) \times [\log(\sum\limits_{o'}L^n(w,o'|m)) - log(P(w|o))]\\
= \sum\limits_w P(w|o) \times \log(\sum\limits_{o'}L^n(w,o'|m)) - \sum\limits_w P(w|o) \times \log(P(w|o))
\end{align*}

Note that that the second term, $- \sum\limits_w P(w|o) \times \log(P(w|o))$, does not depend on the message $m$. For this reason it can be dropped: dropping this term amounts to adding a constant term to the utility of each message (relative to a fixed observation o), which has no effect when we apply the \emph{SoftMax} function in order to derive the speaker's behavior. So we can as well use the following utility function, with no change whatsoever in the behavior of the model:

$$U^n(m,o) = \sum\limits_w P(w|o) \times \log(\sum\limits_{o'}L^n(w,o'|m))$$

Now, we can also consider a model whose general architecture is like ours, where the literal listener $L^0$, in particular, is exactly the same as the one we defined, but where we use the standard utility function of the RSA framework, which is based on the KL-divergence between the joint distribution over $(w,o)$ of the level-$n$ listener, and the joint distribution of the speaker which results from an observation $o$ (such a joint distribution assigns probability 0 to all pairs $(w,o')$ where $o' \neq o$, i.e. $P(w,o'|o) = P(w|o)$ if $o'=o$, otherwise $P(w,o'|o)=0$).

This amounts to moving to the following utility function, which is the standard one in the RSA framework (ignoring the cost term):

$$U^n(m,o) = \sum\limits_w P(w|o) \times \log(L^n(w,o|m))$$

Keeping all the other ingredients of the model presented in sections \ref{RSAmodel} and \ref{concrete}, we obtain, with such a model, numerically different results from those of our main model, but qualitatively similar ones, in the following sense: the pragmatic speaker  (at different recursive depths) can have a preference for an ``around''-statement over any ``between''-statement in some situations where she is able to exclude the peripheral values $0$ and $8$ (and so could have said, e.g., \emph{between 1 and 7}) but has a private distribution that is strongly biased towards values closer to $4$. Crucially, for this model, the limitation result proved in Appendix \ref{app:LU} for the standard LU model does not hold.

\subsection{A variant of the LU model where the utility function is as in our own model} \label{sec:variantLU}

We notate $\sem{m}^i(w)$  the truth-value of the literal meaning of a message $m$, relative to the interpretation function $i$, in world $w$, and $\sem{m}^i$ denotes the set of worlds where $m$ is true relative to interpretation $i$  (in our setting $i$ is what determined the interpretation of `around', i.e. a certain value for $y$). The temperature parameter $\lambda$ is a non-null, positive real number. For any message $m$, $c(m)$ is the \emph{cost} of $m$, a null or positive real number. $P$ is the prior joint distribution on worlds and observations.\\
\\
Below we present the equations that characterize the modified LU model. The crucial difference with the standard LU model shows up in the utility functions (lines 2 and 5), where $L^0(w,o|m,i)$ and $L^n(w,o|m)$ have been replaced, respectively, with $L^0(w|m,i)$ and $L^n(w|m,i)$, which are themselves equal, respectively, to $\sum\limits_{o'}L^0(w,o'|m,i)$ and  $\sum\limits_{o'}L^n(w,o'|m)$.\\
\\
Importantly, the limitation result reported in Appendix \ref{app:LU} for the standard LU model no longer holds for this model.

\begin{enumerate}
	
	\item $L^0(w,o | m,i) = \dfrac{P(w,o) \times \sem{m}^i(w)}{P(\sem{m}^i)}$
	\item $U^1(m|o,i) = (\sum\limits_{w}P(w | o) \times \log(L^0(w | m,i))) - c(m) \\= (\sum\limits_{w}[P(w | o) \times \log(\sum\limits_{o'}L^0(w,o' | m,i))]) - c(m)$
	\item $S^1(m | o, i) = \dfrac{\exp(\lambda U^1(m,o,i))}{ \sum\limits_{m'}\exp(\lambda U^1(m',o,i))}$
	\item $L^1(w,o | m) = \dfrac{P(w,o)\times \sum\limits_{i} P(i).S^1(m | o, i))}{\alpha_1(m)}$, where $\alpha_1(m) = \sum\limits_{w',o'} (P(w',o').\sum\limits_{i} P(i).S^1(m | o', i)))$
	\item For $n \geq 1$, $U^{n+1}(m|o) = (\sum\limits_w P(w| o).\log(L^n(w | m)) - c(m)\\=(\sum\limits_{w}[P(w | o) \times \log(\sum\limits_{o'}L^n(w,o' | m))] - c(m)$
	
	\item $S^{n+1}(m | o) = \dfrac{\exp(\lambda U^{n+1}(m,o))}{\sum\limits_{m'} \exp(\lambda U^{n+1}(m',o))}$
	\item For $n \geq 2$, $L^n(w,o | m) = \dfrac{P(w,o) \times S^n(m | o)}{\alpha_n(m)}$, where $\alpha_n(m) = \sum\limits_{w',o'}P(w',o').S^n(m | o')$

\end{enumerate}

\end{appendices}

\newpage

\bibliographystyle{plainnat}
\bibliography{biblio}

\newpage


\paragraph{Declaration of contribution:} PE and BS share first authorship in this paper. PE came up with the core idea of the paper, which he then developed with SV, BS and AM. PE and SV originally proposed the model in Appendix \ref{app:original}, and PE and BS later worked out its Bayesian counterpart. BS, PE and AM identified the ratio inequality, and AM investigated the model by PE-SV. BS is responsible for the speaker model and the RSA implementation, and for the comparison with the Lexical Uncertainy model. PE and BS jointly wrote the first complete draft of the paper, which the four authors finalized and revised together.

\end{document}